\documentclass[journal]{IEEEtran}

\usepackage{cite}
\usepackage[pagebackref,breaklinks,colorlinks]{hyperref}

\ifCLASSINFOpdf
\else
\fi

\usepackage{amsmath}
\usepackage{amssymb}
\usepackage{booktabs}
\usepackage{multirow}

\usepackage{tikz}
\usepackage{subcaption}
\usepackage{url}

\def\eg{\emph{e.g}\onedot} 
\def\ie{\emph{i.e}\onedot} 
\def\cf{\emph{c.f}\onedot} 
\def\etc{\emph{etc}\onedot} 
 
\def\etal{\emph{et al}\onedot}

\newcommand{\abbr}{DeepTerRa}
\newcommand{\mysubsection}[1]{\textbf{#1}\quad}
\newcommand{\review}[1]{{\color{black}#1}} %
\newcommand{\rebuttal}[1]{{\color{black}#1}} %

\graphicspath{{./Images/}{./Figures/result/}{./Images/result/}}

\begin{document}
\title{Learning Digital Terrain Models from Point Clouds: ALS2DTM Dataset and Rasterization-based GAN}

\author{Hoàng-Ân Lê,
    Florent Guiotte,
    Minh-Tan Pham,
    Sébastien Lefèvre, and
    Thomas Corpetti

\thanks{H.-Â. Lê, F. Guiotte, M.T. Pham and S. Lef\`evre are with the Institut de recherche en informatique et systèmes aléatoires (IRISA), UMR 6074, Université Bretagne Sud, 56000 Vannes, France. H.-Â. Lê is also with the France Energies Marines, 29200 Brest, France.} 
\thanks{T. Corpetti is with the Littoral - Environnement - Télédétection - Géomatique (LETG), UMR 6554, Université Rennes 2, 35000 Rennes, France.} 
\thanks{Corresponding e-mail: hoang-an.le@irisa.fr; minh-tan.pham@irisa.fr, and sebastien.lefevre@irisa.fr}

}

\markboth{Submitted to Journal of Selected Topics in Applied Earth Observations and Remote Sensing }%
{Lê \MakeLowercase{\textit{et al.}}: Bare Demo of IEEEtran.cls for IEEE Journals}

\maketitle

\begin{abstract}
    Despite the popularity of deep neural networks in various domains,
the extraction of
digital terrain models (DTMs) from airborne laser scanning (ALS) point clouds is still challenging.
This might be due to the lack of dedicated large-scale
annotated dataset and the data-structure discrepancy between point clouds and DTMs.
To promote data-driven DTM extraction, this paper collects from open sources a large-scale dataset of ALS point clouds and corresponding
DTMs with various urban, forested, and mountainous scenes.
A baseline method is proposed
as the first attempt to train a \underline{Deep} neural network to extract digital \underline{Ter}rain models directly from ALS point clouds via \underline{Ra}sterization techniques, coined DeepTerRa.
Extensive studies with well-established methods are performed to benchmark the 
dataset and analyze the challenges in learning to extract DTM from point clouds.
The experimental results show the interest of the 
agnostic data-driven approach, with sub-metric error level compared to 
methods designed for DTM extraction. The data and source code is provided at \url{https://lhoangan.github.io/deepterra/} for reproducibility and further similar research.

\end{abstract}


\IEEEpeerreviewmaketitle

\begin{figure*}[t]
    \centering
    \includegraphics[width=.19\textwidth]{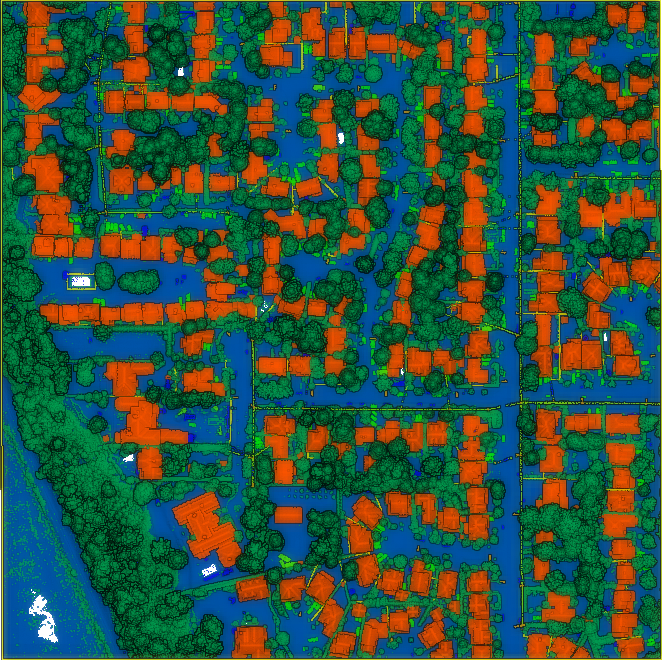}
    \includegraphics[width=.19\textwidth]{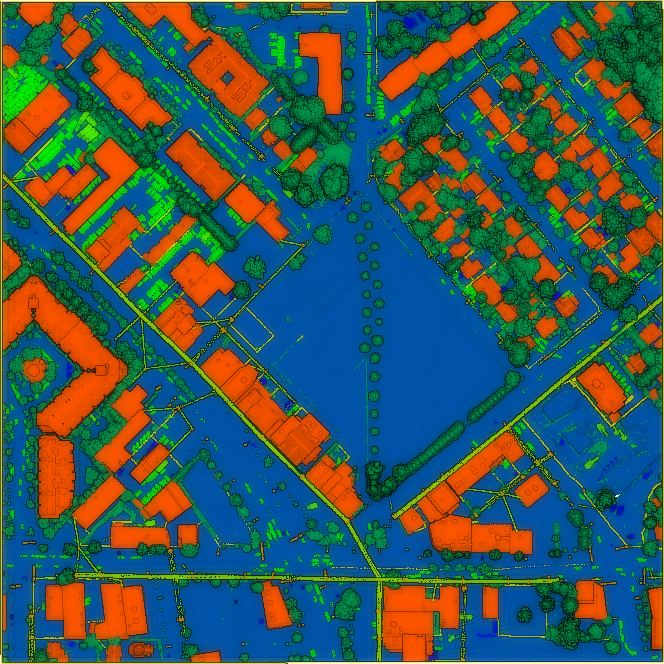}
    \includegraphics[width=.19\textwidth]{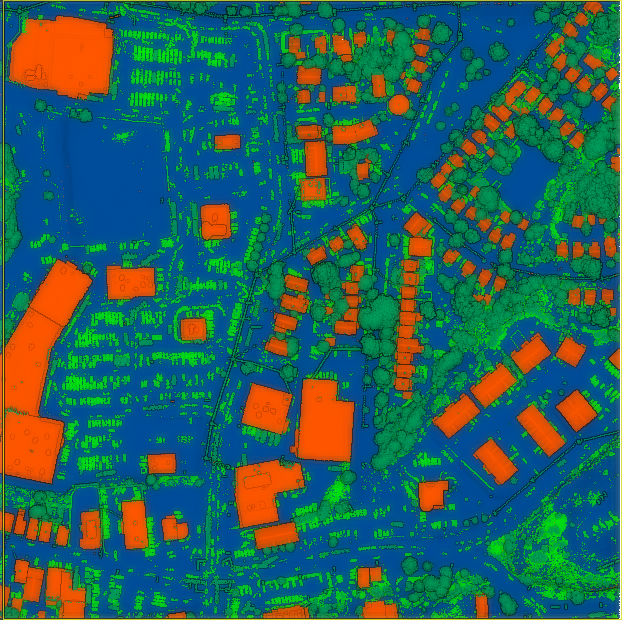}
    \includegraphics[width=.19\textwidth]{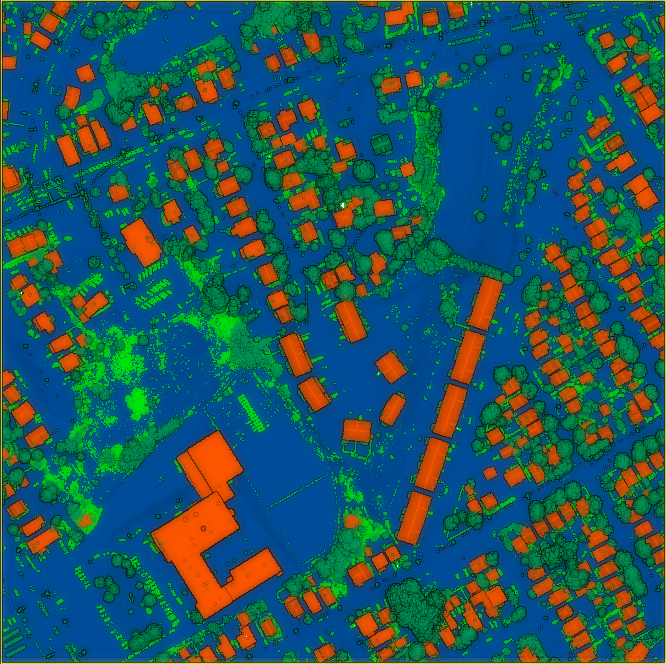}
    \includegraphics[width=.19\textwidth]{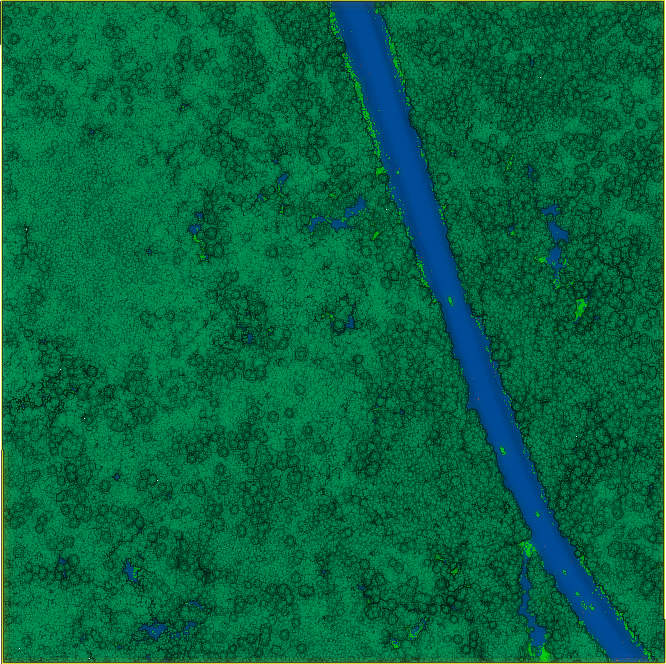} \\
    \includegraphics[width=.19\textwidth]{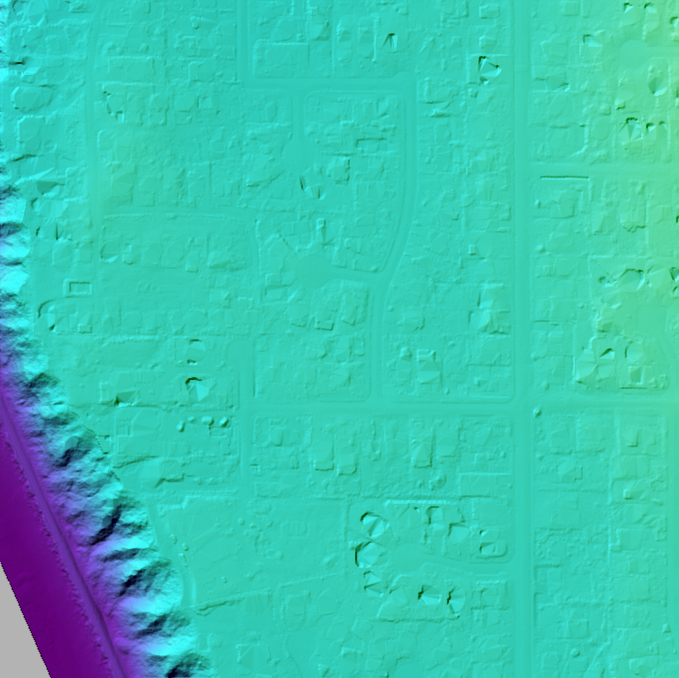}
    \includegraphics[width=.19\textwidth]{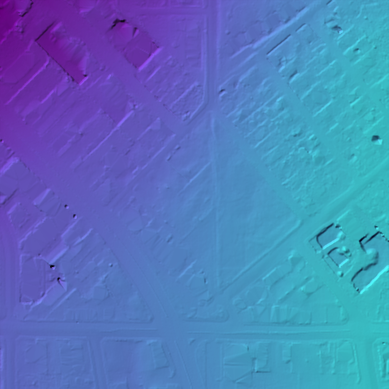}
    \includegraphics[width=.19\textwidth]{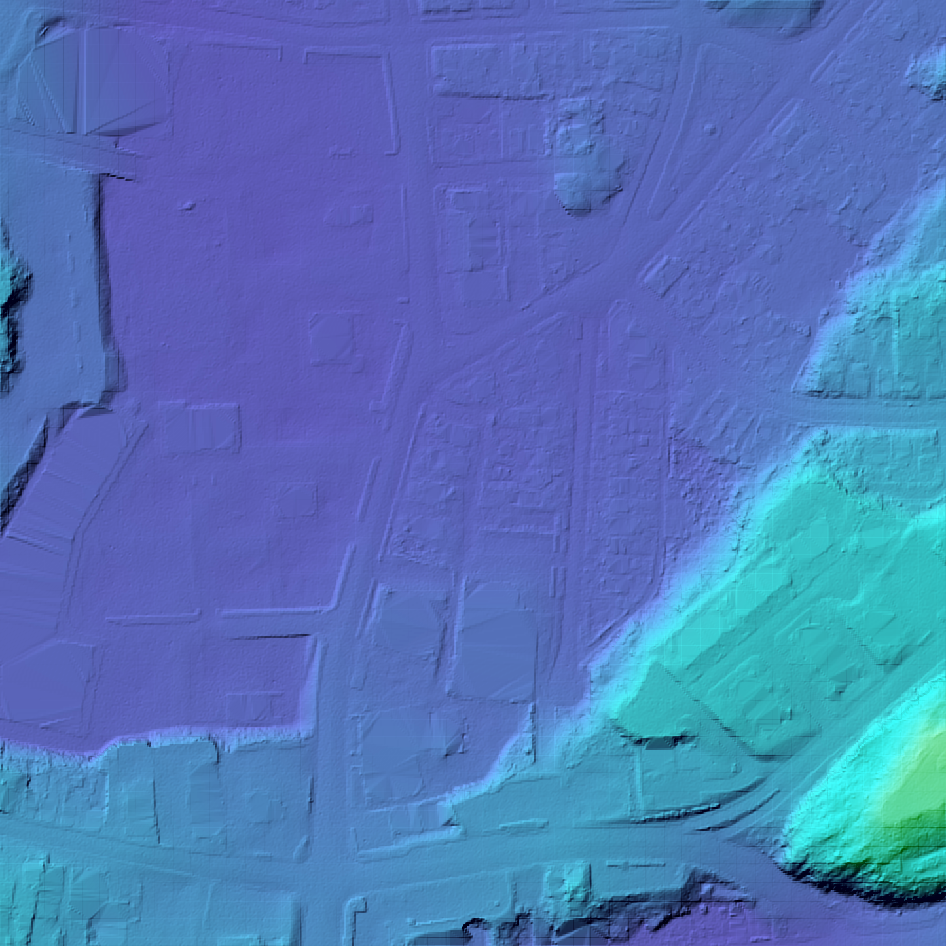}
    \includegraphics[width=.19\textwidth]{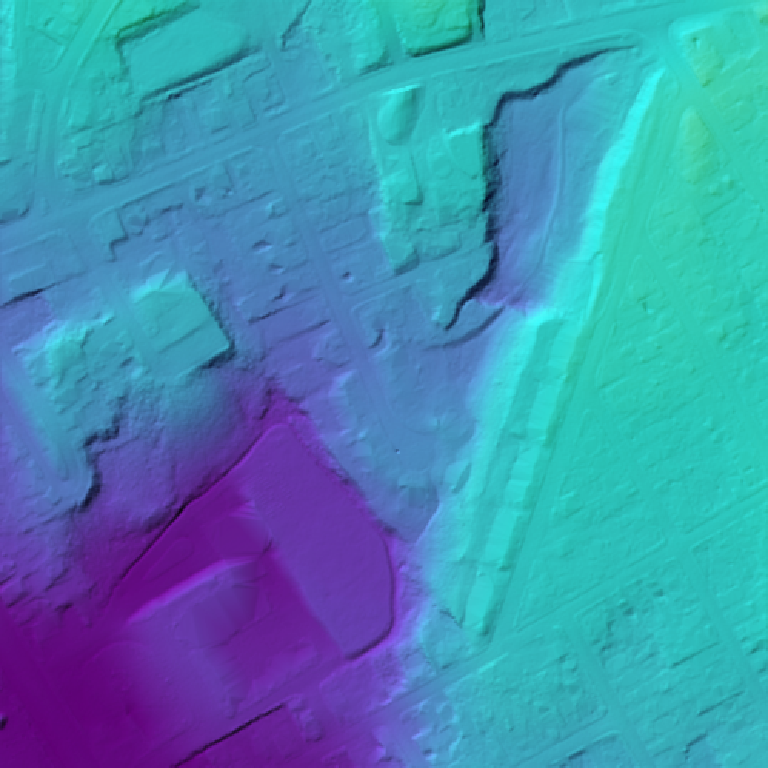}
    \includegraphics[width=.19\textwidth]{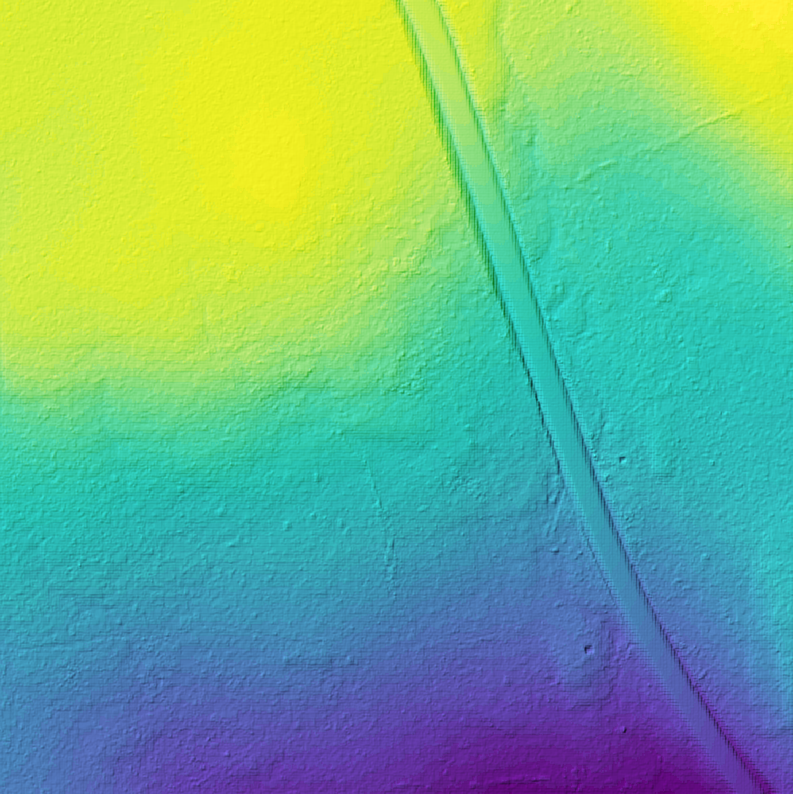}\\
    \caption{
    \abbr\ dataset with Aerial Laser Scanning Points Clouds (top) and reference Digital Terrain Models (bottom).
    }
    \label{fig:teaser}
\end{figure*}

\section{Introduction}
\label{sec:Intro}

Digital terrain models (DTMs), 2D representations of bare-earth
surface elevation, are critical for geographical and environmental
studies such as flood modelling~\cite{Bakula2011,Muhadi2020,Muthusamy2021},
above-ground biomass estimation~\cite{Gonzalez2019}, infrastructure
planning~\cite{Xu2021},~\etc. 
Being a subset of digital elevation models (DEMs),
DTMs are usually distinguished from its relative, digital surface models (DSMs), by the
exclusion of objects (houses, vegetation,~\etc) and representing only the bare 
ground surface underlying the objects.

DTM extraction can be achieved from satellite or aerial
images~\cite{Gevaert2018, Rokhmana2019}, DSMs~\cite{Duan2019}, or point 
clouds~\cite{Debella2016,Ye2019}. Among the possibilities, airborne laser
scanning (ALS) point clouds, with rich geometry information (\cf~images) and
high precision due to the penetration capability through vegetation
(\cf~photogrammetry), have gained attention in the recent decades. Several studies
were proposed on ALS point clouds, while datasets were introduced
to accommodate data-driven deep learning-based research, notably for
point cloud segmentation~\cite{dublin2019,dales2020,lasdu2020,opengf2021}.

Deep learning methods, which learn to identify salient patterns from high volume
of data, have proven to be more effective over well-engineered features
for several computer vision and machine learning problems~\cite{Krizhevsky2012,Long2015,rcnn}.
For DTM extraction, different CNN-based methods have been proposed for point cloud ground
filtering~\cite{Hu2016,Zhang2020,opengf2021}, one of the two elemental steps,
while the elevation interpolation problem is left open.

One hindrance for data-driven DTM extraction is the lack of dedicated
large-scale datasets. While ground filtering can be formulated as a semantic
segmentation problem, where one class is ground and the rest are non-ground, creating
ground truth DTMs requires specialized knowledge and tools.
To encourage deep-learning-based research in this important field,
we introduce a large-scale dataset of ALS point clouds and reference DTMs
collected from open data sources (Figure~\ref{fig:teaser}).
Several well-established methods are compared for benchmarking purposes.

Another challenge to be tackled is the discrepancy of inputs and outputs representations.
3D ALS point clouds inherently lack topological information that 2D image-like DTMs embrace.
In this paper, we propose using rasterization techniques to bridge the
representation gap, which has proven to largely retain point-cloud
information~\cite{Guiotte2020}. As such, point-cloud-based DTM extraction can be formulated
as an image-to-image translation problem~\cite{Isola2017} and thus off-the-shelf computer
vision methods can be applied.
As rasterization is the idea behind rendering
networks,~\eg~RenderNet~\cite{Nguyen2018}, it could be extended to an end-to-end solution.

To this end, we endeavor to predict DTM in a unified deep learning framework
by concurrently tackle both the elemental problems, ground filtering and elevation interpolation.
The proposed baseline employs an off-the-shelf architecture and generative adversarial network
(GAN)~\cite{GAN} to predict DTM from point-cloud-based multi-channel
rasters. The experimental results show that simple employment of computer vision
techniques approaches well-established methods specifically designed for DTM
extraction. Extensive analyses are performed to identify the shortcomings and
potentials of our data-driven approach for DTM extraction.

The main contributions of this paper are thus as follows:
\begin{itemize}
    \item the first large-scale dataset combining ALS point cloud and bare-earth elevation (DTM),
     covering 52 km$^2$ with several well-established methods for benchmarking;
    \item  the first attempt, to the best of our knowledge, to formulate the DTM
    extraction problem in a unified deep learning framework that could
    potentially be developed into an end-to-end solution;
    \item a comparative study of different rasterization strategies for 
    DTM extraction from ALS point clouds.
\end{itemize}

\section{Related Work}
\label{sec:related_work}

\subsection{DTM extraction}
\label{subsec:dtm_extraction}

DTM extraction is a long-standing problem in geospatial processing. It reconstructs
from remote sensing data the bare-earth surface underlying objects, or land covers,
both natural (trees, water,~\etc) and artificial (buildings, construction,~\etc).
The typical process composes of 2 steps, ground filtering and surface
interpolation~\cite{Duan2019,Ye2019}.
Ground filtering, or terrain extraction, identifies the ground portions from
among other entities (water, trees, buildings,~\etc) and extracts them
from the input geospatial data. The object-free ground portions then have their
surface completed by interpolation methods such as regularized splines~\cite{Hu2016},
elevation gridding~\cite{Hutchinson2011}, or using affine-kernel
embeddings~\cite{Duan2019}.

Due to high topographical variation and complex structures of terrain types
(\eg~flat, valley, hill, mountainous) and objects (\eg~vegetation, construction)
in geospatial data, efforts have been focused on exploring ground filtering.
One strategy is to rely on terrain elevation and
slope to establish rules that separate objects from
ground~\cite{Debella2016,Gevaert2018,Sithole2001,Sithole2004}, under the
assumption of gradual relief and small land covers with respect to
land scales. Classification is performed by thresholding over pre-defined
values either by chosen manually or from local adaptive filtering based on terrain
slope~\cite{Sithole2001}.

Another approach is to approximate the terrain surface and identify objects
using neighboring information.
The progressive morphological filter (PMF)~\cite{Zhang2003} uses sliding windows
of increasing size to filter the points with elevation difference.
Similarly, the simple morphological filter (SMRF)~\cite{Pingel2013} filters the
points with linearly increasing windows and slope thresholding.
Such methods still rely on thresholds selection for the size of windows and the
elevation functions. One threshold-free alternative named skewness balancing method 
(SBM)~\cite{Bartels2010} is based on unsupervised statistical analysis of point
cloud skewness to filter ground points.
The most recent cloth simulation filtering (CSF)~\cite{Zhang2016} is based on
cloth simulation to filter ground points by tuning the resolution of the virtual
cloth, its rigidness, and gravity parameters.
Despite being simple and computationally efficient,
choosing optimal threshold remains the hindrance for this strategy.

\begin{figure*}[t]
    \centering
    \begin{tikzpicture}
        \node at (0, 0) {
            \includegraphics[width=.12\textwidth]{Images/DALES-5110-54495.png}
            \includegraphics[width=.12\textwidth]{Images/5110_54495-dtm.png}
            \includegraphics[width=.12\textwidth]{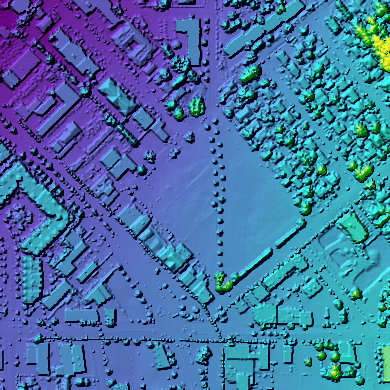}
            \includegraphics[width=.12\textwidth]{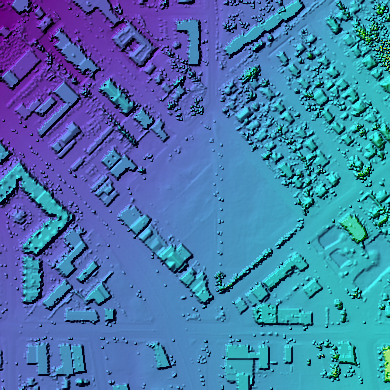}
            \includegraphics[width=.12\textwidth]{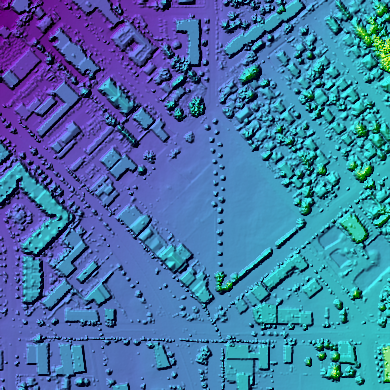}
            \includegraphics[width=.12\textwidth]{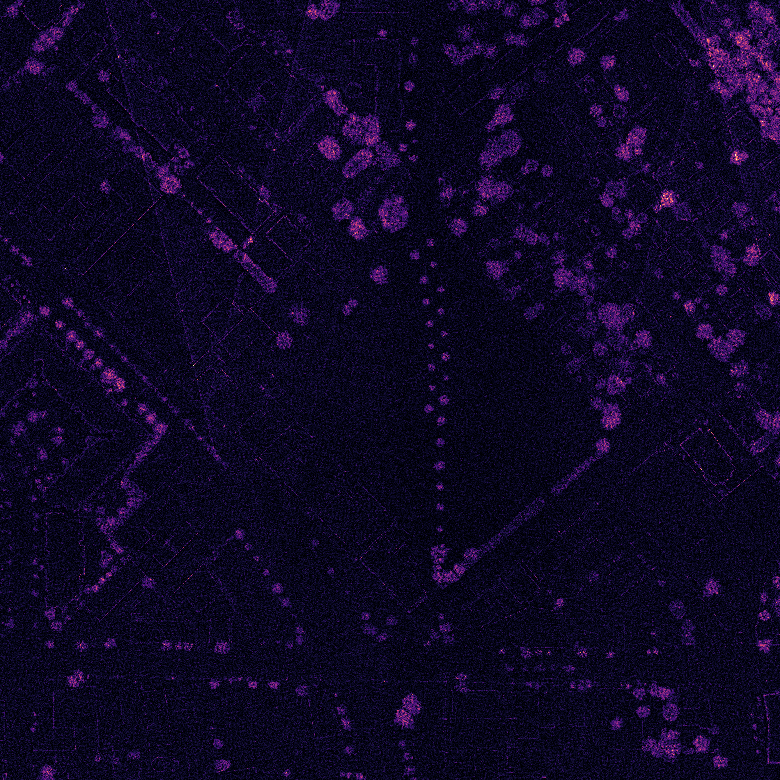}
            \includegraphics[width=.12\textwidth]{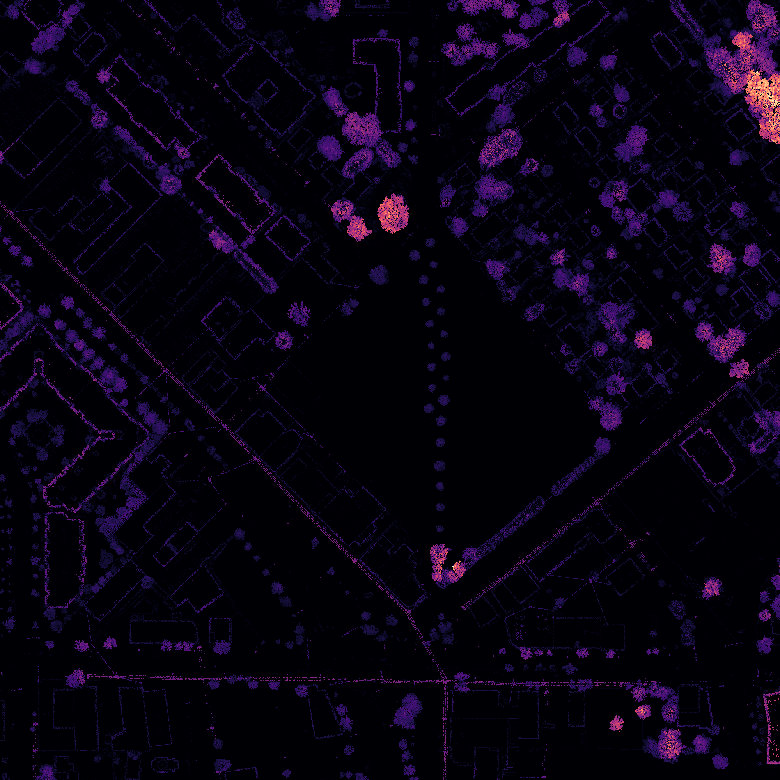}
            \includegraphics[width=.12\textwidth]{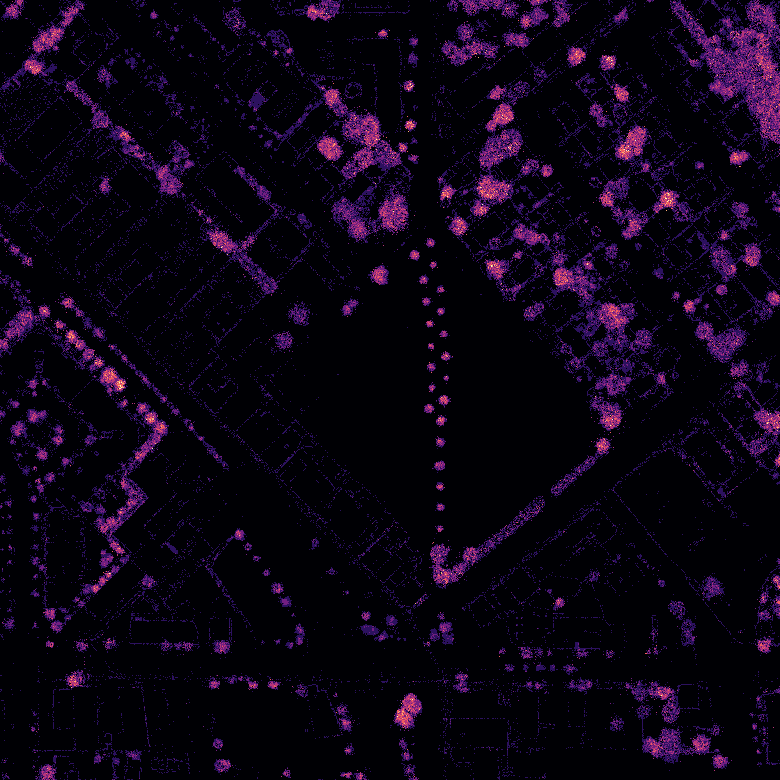}
        };
        \node at (0, -2.2) {
            \includegraphics[width=.12\textwidth]{Images/NB-25360-73650.png}
            \includegraphics[width=.12\textwidth]{Images/25360_73650-dtm.png}
            \includegraphics[width=.12\textwidth]{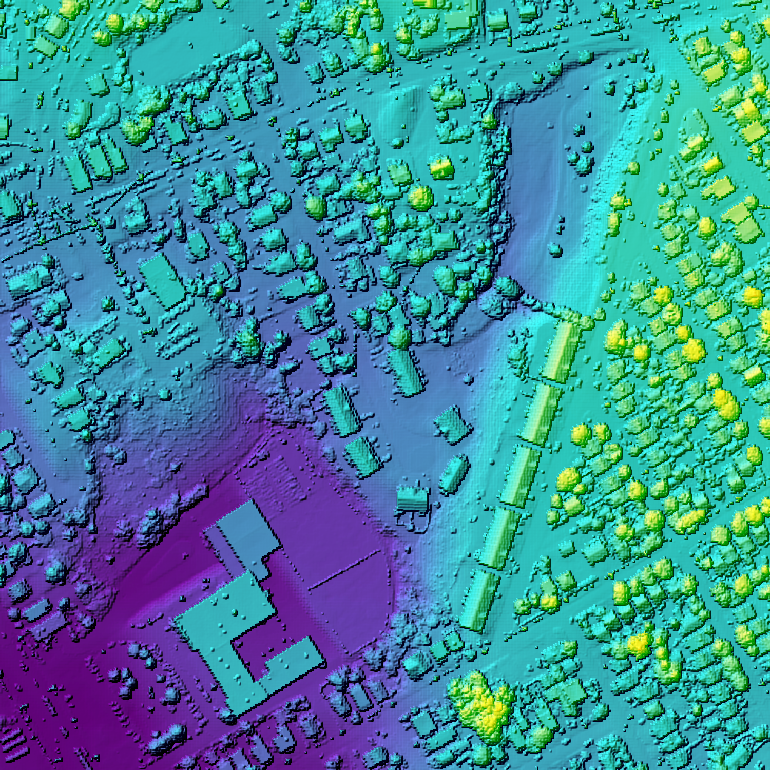}
            \includegraphics[width=.12\textwidth]{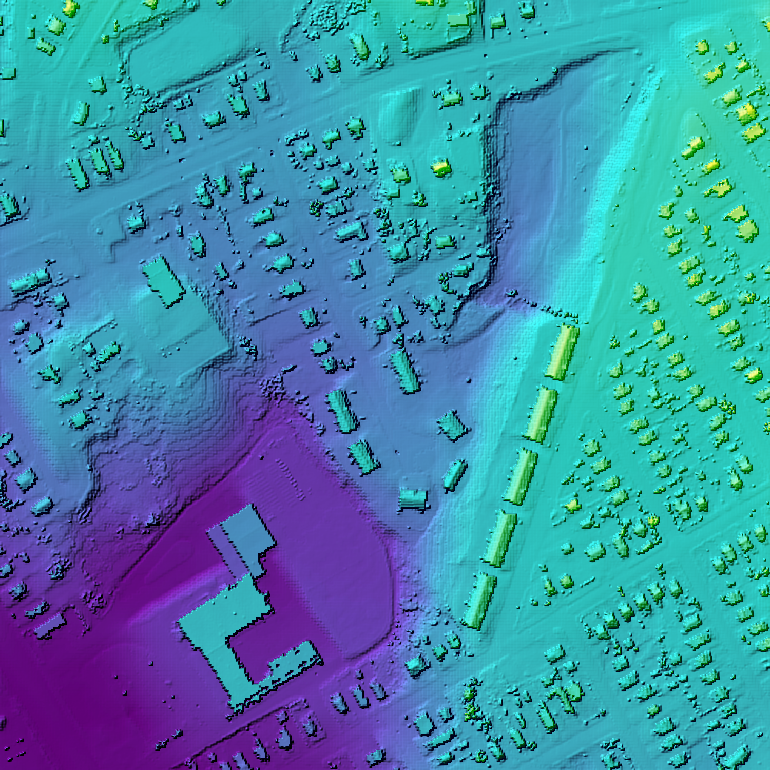}
            \includegraphics[width=.12\textwidth]{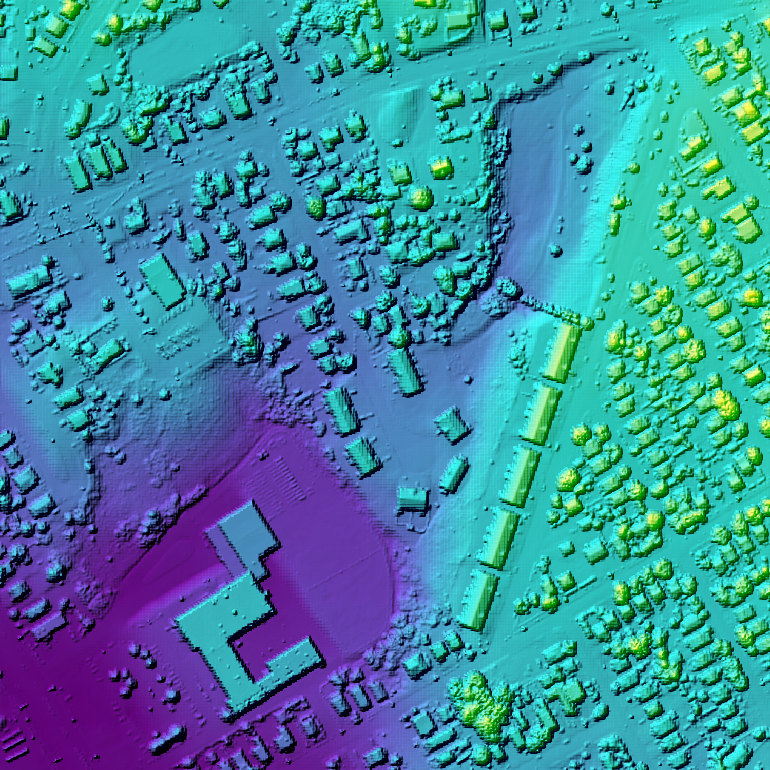}
            \includegraphics[width=.12\textwidth]{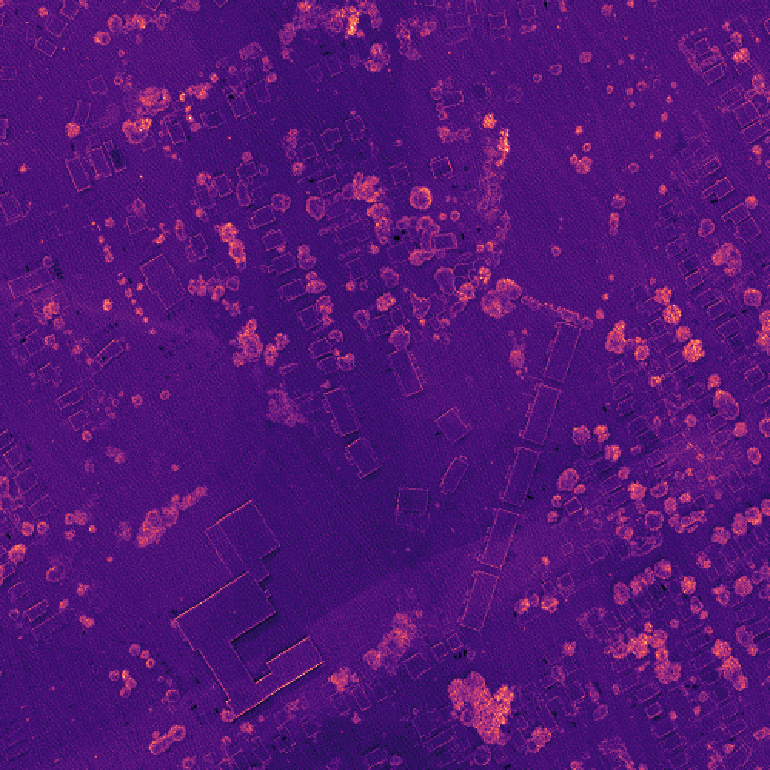}
            \includegraphics[width=.12\textwidth]{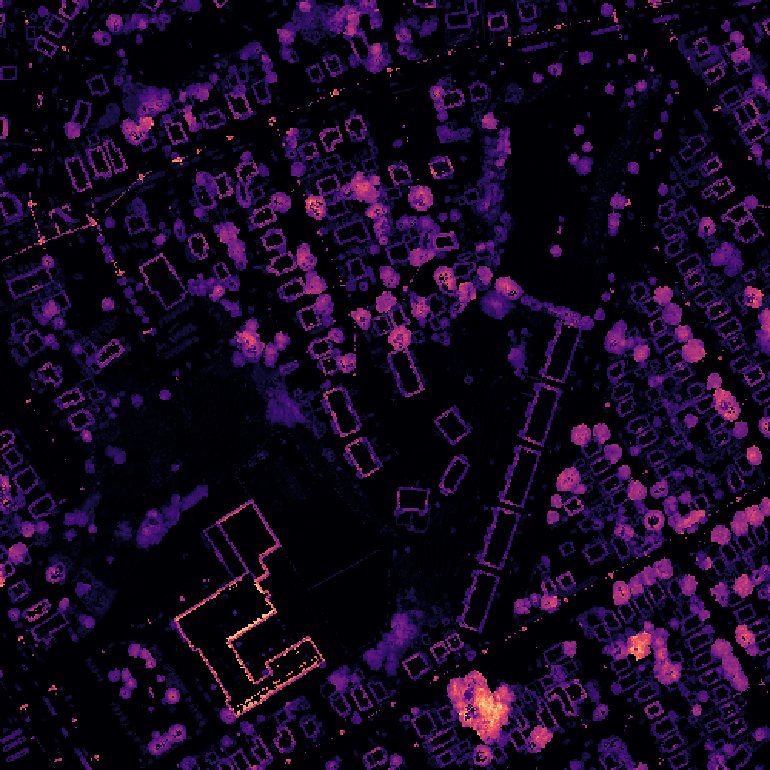}
            \includegraphics[width=.12\textwidth]{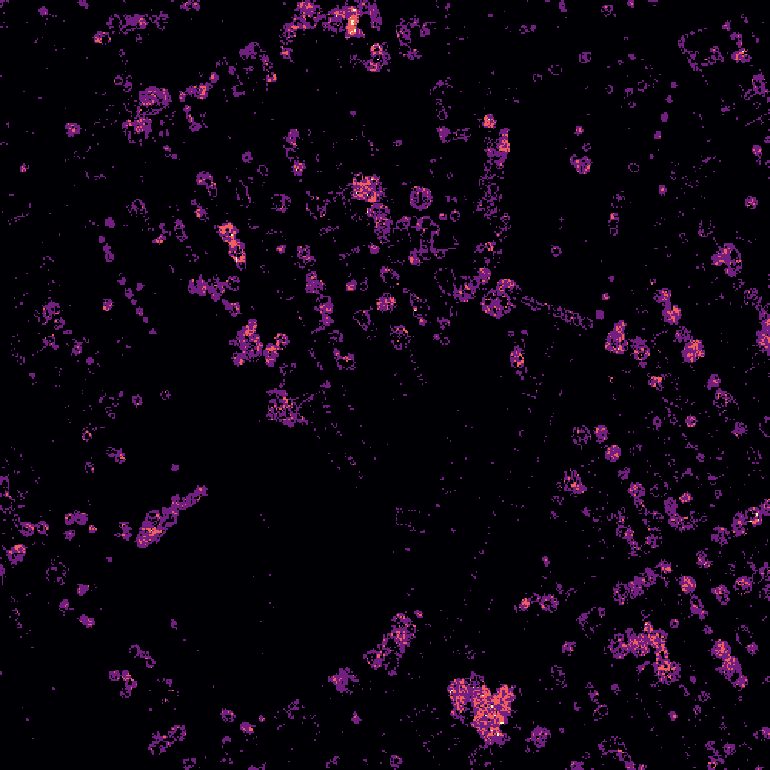}
        };
        \node at (-8.0,1.3) {\small Point cloud (top)};
        \node at (-5.7,1.31) {\small DTM};
        \node at (-3.4,1.3) {\small voxel-top};
        \node at (-1.1,1.32) {\small voxel-bottom};
        \node at ( 1.1,1.3) {\small pixel-mean};
        \node at ( 3.4,1.3) {\small density};
        \node at ( 5.7,1.31) {\small stdev};
        \node at ( 8.0,1.3) {\small echoes};
    \end{tikzpicture}
    \caption{Visualization of different rasters used in the paper for an exemplar of DALES (\emph{top}) and NB (\emph{bottom}) with nadir-view point cloud (\emph{first column}),
    and reference DTM (\emph{second column}). Voxel-top is commonly known as DSM.}
    \label{fig:raster}
\end{figure*}

To overcome window size sensitivity and reduce parameterization,
Duan~\etal~\cite{Duan2019} exploit DSMs through a combination of rule-based classification
and multiscale morphological analysis and define several different thresholds
based on input DSM statistics. The parameters are said to be robust to various
DSM types with arbitrary quality and not require tuning.
Another DSM-based DTM extraction aims to preserve
sharp terrain edges by using anisotropic filtering~\cite{Pijl2020}.

In recent years, the paradigm is shifted toward more of 
learning-based methods~\cite{Gevaert2018,Hu2016,opengf2021,Zhang2020}.
Such methods find their emergence from the breakthroughs of deep
networks~\cite{imagenet,He2015,Krizhevsky2012,Long2015}, which, 
inspired by human neural systems, model and learn parametric nonlinear
functions defined by sequentially stacked layers (\textit{hence} deep network) 
from massive amount of data. The convolutional neural network (CNN) has proven
to be effective for various computer vision tasks, achieving 
state-of-the-art results from high-level tasks (\eg~classification~\cite{Krizhevsky2012,He2015} and segmentation~\cite{Long2015}),
to low-level (\eg~optical flow~\cite{Sun2018} and surface normal estimation~\cite{HALe2018}).

Benefiting from the advancement of computer vision, several research studies
have tried to apply deep learning techniques to DTM-related problems.
\cite{Tapper2016,Gevaert2018} extract DTMs from satellite imagery using
fully convolutional network (FCN)~\cite{Long2015} and residual network 
(ResNet)~\cite{resnet} architectures.
Ground filtering has also received special attention:
\cite{Hu2016,Ye2019} extract ground from ALS point clouds
by first sampling into image patches before performing image-based classification,
while~\cite{Zhang2020} applies dynamic graph convolution
techniques~\cite{Wang2019} and works directly on point clouds to benefit from 
the geometric structures.

\review{
On the other hand,~\cite{Ayhan2020,Kwan20,Crema2020} assume the availability of
land cover classes, i.e. from ground filtering, and compare 
various image inpainting methods, among which are the GAN-based~\cite{Yu2018},
in completing the DTMs.
Generative adversarial networks (GAN)~\cite{GAN} employ adversarial losses,
usually presented as a zero-sum game between 2 networks known as generators and
discriminators in which one's gain is the other's loss, to improve quality of
generated samples. The framework plays as the basis for a wide range of
application such as image-to-image translation~\cite{Isola2017},
image inpainting~\cite{Yu2018}, or downstream tasks as DTM extraction from
RGB images~\cite{Kwan20,Panagiotou2020}.
Different from our approach, these methods either assume imagery inputs
or only focus on a single step in DTM extraction. Our method may share
similar traits to an inpainting method for GAN-based constructing DTMs,
the main differences are the use of rasterization to accentuate
various point cloud features and indirect use of semantic information
to examine its benefit.
}
To the best of our knowledge,
there is hardly any deep learning-based method that attempts to extract DTM from
ALS point clouds  data directly.
This could be due to the discrepancy between topologically unordered
point cloud inputs and the well organized image-like DTMs, \review{which is one
of the targets this paper seeks to tackle.}

\subsection{Datasets}

Another explanation for the limited deep-learning-based DTM extraction
from ALS point clouds is the current shortage of dedicated large-scale labelled 
datasets. Despite the effectiveness and popularity of deep learning techniques,
large-scale datasets play a critical role in their success. 
Existing datasets with ALS point clouds include
Filtertest\footnote{\url{https://www.itc.nl/isprs/wgIII-3/filtertest}},
ISPRS2012~\cite{isprs2012},
DublinCity~\cite{dublin2019},
LASDU~\cite{lasdu2020}, and
DALES~\cite{dales2020}.
However, they are all designed for urban semantic segmentation tasks
(with the exception of Filtertest which is small reference data for ground
filtering), thus not equipped with DTM data \rebuttal{(Table~\ref{tab:compare})}.

ALS point cloud datasets for deep-learning-based ground filtering
were used in~\cite{Hu2016,Zhang2020}, yet the data are either
not publicly released or limited to single scene types.
The most recent OpenGF dataset~\cite{opengf2021}, collected from
open data sources, also targets land-level ground filtering
only, and does not include elevation information.
\rebuttal{
Other related work including~\cite{Zhu2018,Han2021} which model
large scale urban scenes from MVS Meshes.
}

In this work, we collect and introduce a large-scale dataset, coined \abbr,
including ALS point clouds (1,659M points, covering 52 km$^2$) with semantics
and reference DTM data. To assist ongoing research, part of the dataset is 
augmenting the DALES~\cite{dales2020} dataset with elevation data, while the
rest are from different regions. To the best of our knowledge, this is the
currently first and largest dataset with ALS point clouds with corresponding
DTM data, to encourage deep learning research in DTM extraction and remote
sensing.

\begin{table}[t]
    \centering
    \setlength{\tabcolsep}{2pt}
    \begin{tabular}{@{}lcccccc@{}}
        \toprule
        Dataset  &  year & \#points & coverage & DTM & semantics \\
        \midrule
        Filtertest & 2003  & 2.3M & 1.1 km$^2$ &  $\checkmark$ & 2 classes \\
        ISPRS2012~\cite{isprs2012} & 2012  & 1.2M & & & 9 classes \\ %
        DublinCity~\cite{dublin2019} & 2019 & 260M & 2 km$^2$ & & 13 classes \\
        LASDU~\cite{lasdu2020} & 2020 & 3.1M & 1 km$^2$ & & 5 classes \\
        DALES~\cite{dales2020} & 2020 & 505M & 10 km$^2$ & & 8 classes \\
        OpenGF~\cite{opengf2021} & 2021 & 542.1M & 47.7 km$^2$ & & 3 classes \\
        \midrule
        DALES (ours) & 2020 & 505M & 10 km$^2$ &  $\checkmark$ & 8 classes \\
        NB (ours)    & 2022 & 1153M & 42 km$^2$ & $\checkmark$ & 11 classes \\
        \bottomrule
    \end{tabular}
   \caption{\rebuttal{Comparison of the dataset proposed in our paper and existing ones reported the literature. Key features of our dataset include reference  DTM information for ALS point clouds, with larger coverage and higher number of points.}}
   \label{tab:compare}
\end{table}

\section{\abbr}
\label{sec:method}

\begin{figure*}[t]
    \centering
    \def\svgwidth{\textwidth}
    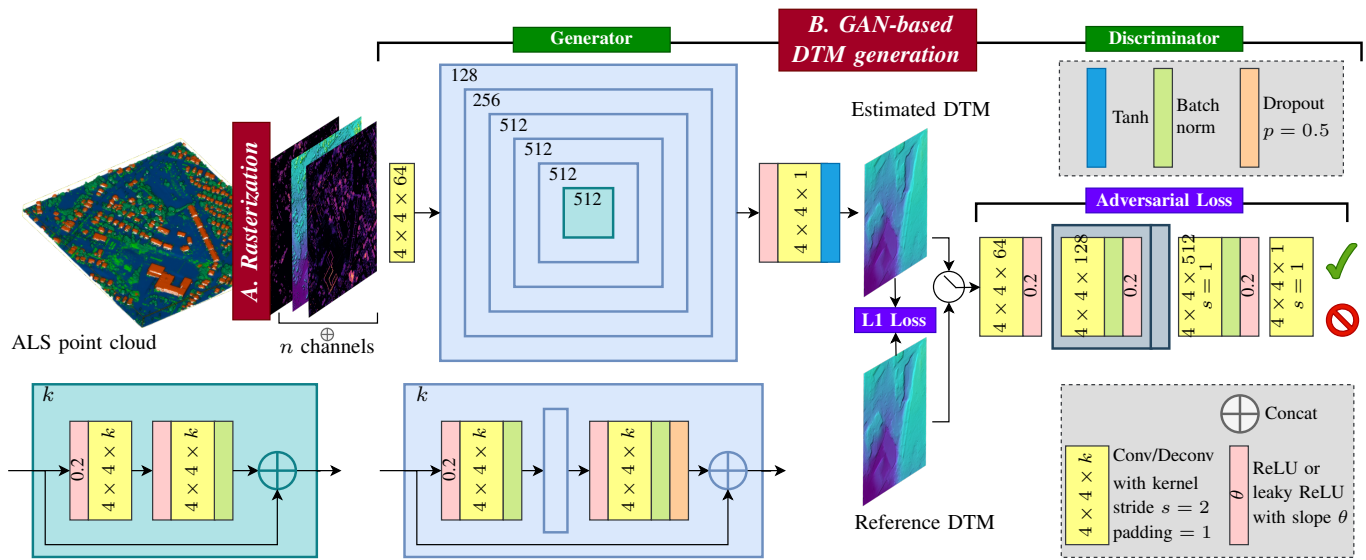
    \caption{\rebuttal{The pipeline overview: an ALS point cloud is rasterized with multiple
    strategies, the resulted rasters are concatenated and fed to a deep neural network
    to generate DTMs. L1 and adversarial losses are used to train the network. 
    \review{ Repeated blocks are color-coded.
    The U-Net generator is shown in a recursive manner with the recursive blocks in blue and
    terminating (base) block in green. Each block with a U-shape connection contains the same structure and number of layers. The only difference is the number of convolutional layers indicated by $k$. (Best viewed in color.)
    }
    }
    }
    \label{fig:pipeline}
\end{figure*}

In this section, we detail the deep-learning-based approach for extracting DTM
from ALS point clouds.
To bridge the topological gap between unordered point cloud inputs and 
image-like DTM targets, we employ and elevate the rasterization technique
to extract information from point clouds. The rasters then serve as
inputs to a CNN-based generator which is trained with an adversarial loss
(discriminator)  and outputs a DTM. The pipeline overview is shown in
Figure~\ref{fig:pipeline}.

\subsection{Rasterization}
\label{subsec:rasterization}

\newcommand{\x}{\boldsymbol{x}}

ALS mechanism allows capturing high scene details by projecting laser beams on
a scene and recording the returning beams. Due to the penetration
ability of laser, a single emitting beam may result in several returns, or echoes,
especially in vegetation areas, by interacting with different objects along the height,
such as foliage, leaves, branches, ground,~\etc. Each echo registers a point
${\bf p} = \{ (x, y, z, I, e,...) \}$ with spatial coordinates $x, y, z$ of
the interacted surface, intensity $I$, echo number $e$,~\etc.

Rasterization arranges an unordered set of points
$P=\{{\bf p}_k\}$ recorded by ALS in a regular 2D grid, or raster.
The raster size is determined by the spatial extension of the points on the ground
plane ($xy-$plane) and the size of each cell into which the points are quantized.
Each grid cell, or raster pixel, defines a vertical column in the 3D space, which can
be further quantized into cubic cells or voxels. In this paper, a fixed size of 25cm cubed
voxels are used.

Depending on the spatial resolution and scene structures, each pixel or voxel may
contain zero, one, or several points. Different rasterization strategies apply
different methods to make use of the multiple-point information.
We propose to derive several rasters to extract most of the information contained
in the ALS data \rebuttal{(Fig.~\ref{fig:raster})}.
In particular, 2 levels of rasterization are considered, namely
\emph{voxel level}, treating all points in a cubic cell at a time, and
\emph{pixel level}, the whole vertical column altogether.

Depending on information being rasterized, the resulted rasters can be categorized
into 2 groups, elevation and statistic rasters.

\mysubsection{Elevation rasters} The vertical coordinate $z$ of the points are rasterized.
Three strategies are proposed, namely (1) \emph{pixel-mean} which registers the mean
of all points in a vertical column, (2) \emph{voxel-top} storing the mean of the
upper points, and (3) \emph{voxel-bottom} for the lower points. As such, the elevation
rasters contain particularly the elevation information of a point cloud,
which can be used to infer DTM. In fact, the voxel-top rasters are also known as
DSMs.

\mysubsection{Statistic rasters} The statistic rasters do not specify 
elevation but the statistics of the scene structures hinting the
object types and elevation. In this paper, we propose 3 raster types, all 
at pixel-level:
(1) the \emph{density} raster provides the number of points at each pixel, which is
useful to assess pixel-wise reliability when processing other rasters;
(2) the standard deviation (\emph{stdev}) raster describes the variation of
the voxels' mean elevation in each column;
(3) the \emph{echoes} raster gives the mode of the echo numbers or the most frequent
echo number in a column, showing how laser beams interact with the objects
encountered at each location. As such, the raster signifies the object
types.~\eg high numbers, or multiple echoes, usually indicate the presence of vegetation.

\review{
As such, each raster is a single-channel image-like object representing
different aspects and spatial information of the point cloud at each pixel.
Hence, different from typical usage of DSM or RGB images for DTM extraction, we do not
limit the usage to just 1 type of rasters at a time. To make the best use of various
information, the generated rasters are concatenated along the depth channel
before being fed to the network. Analyses on the rasters' contributions
are presented in Sec.~\ref{sec:experiments}.
}

\begin{figure*}
    \centering
    \subcaptionbox{Without correction\label{fig:offset:no}}
        {\includegraphics[width=.35\linewidth]{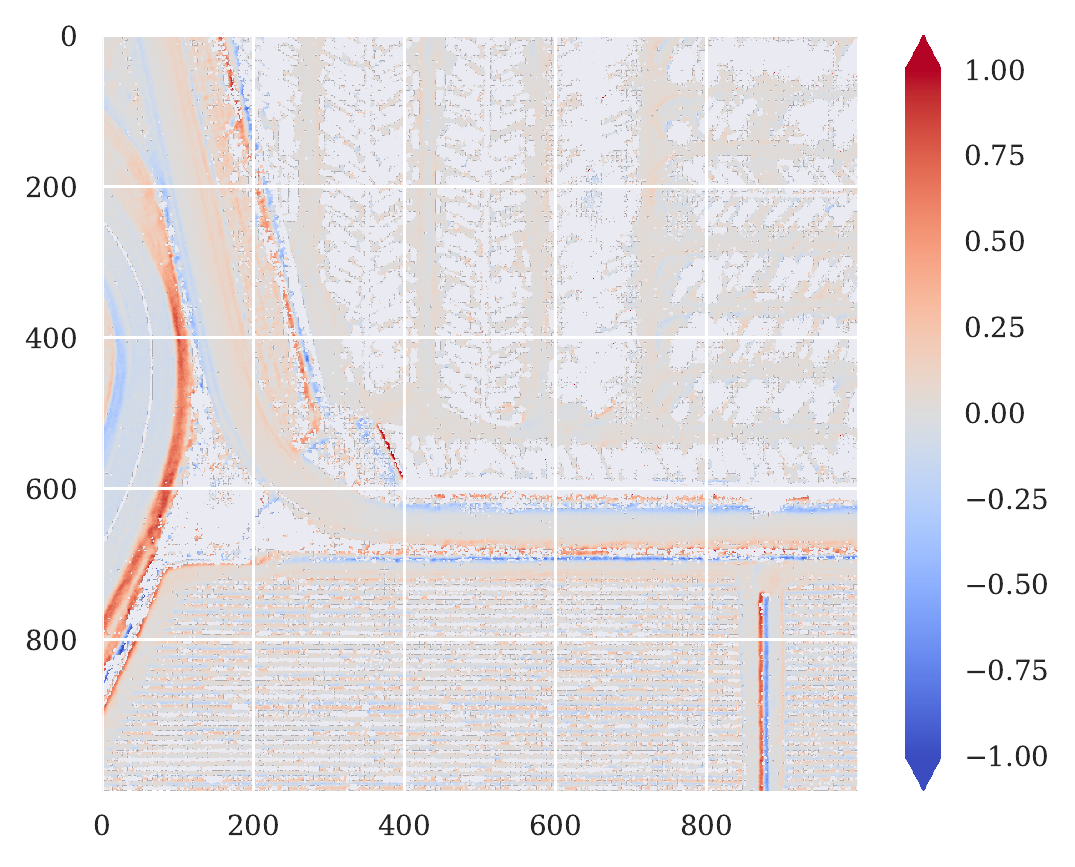}}%
    \subcaptionbox{Greedy grid offset search\label{fig:offset:rmse}}
        {\includegraphics[width=.30\linewidth]{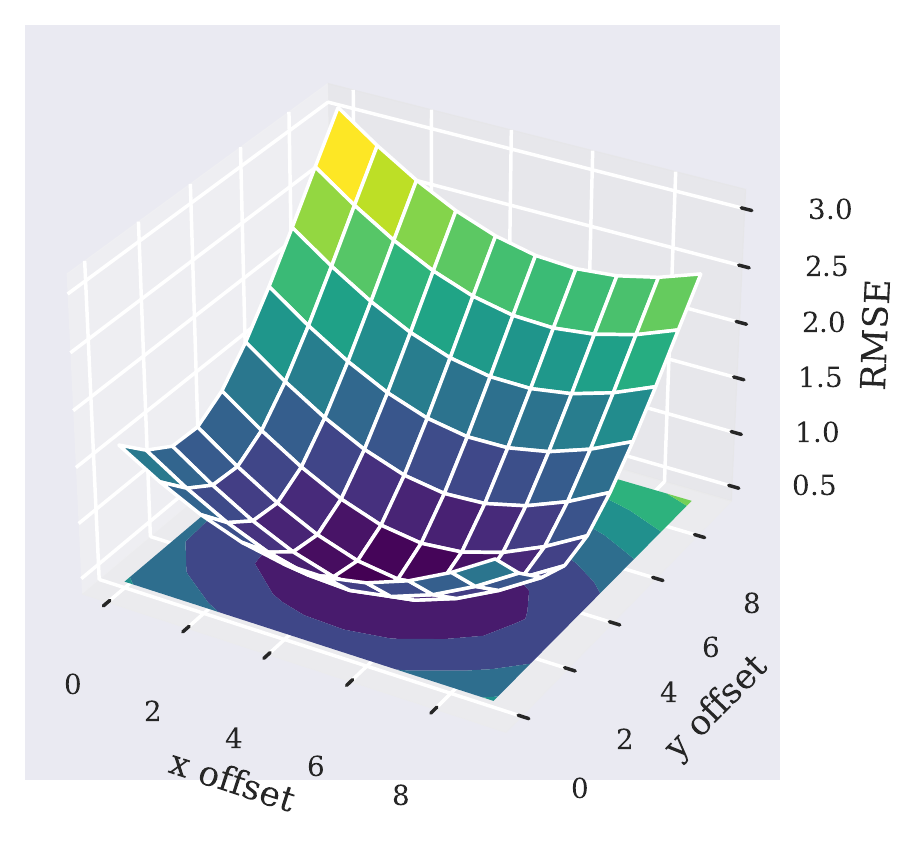}}%
    \subcaptionbox{With correction \label{fig:offset:yes}}
        {\includegraphics[width=.35\linewidth]{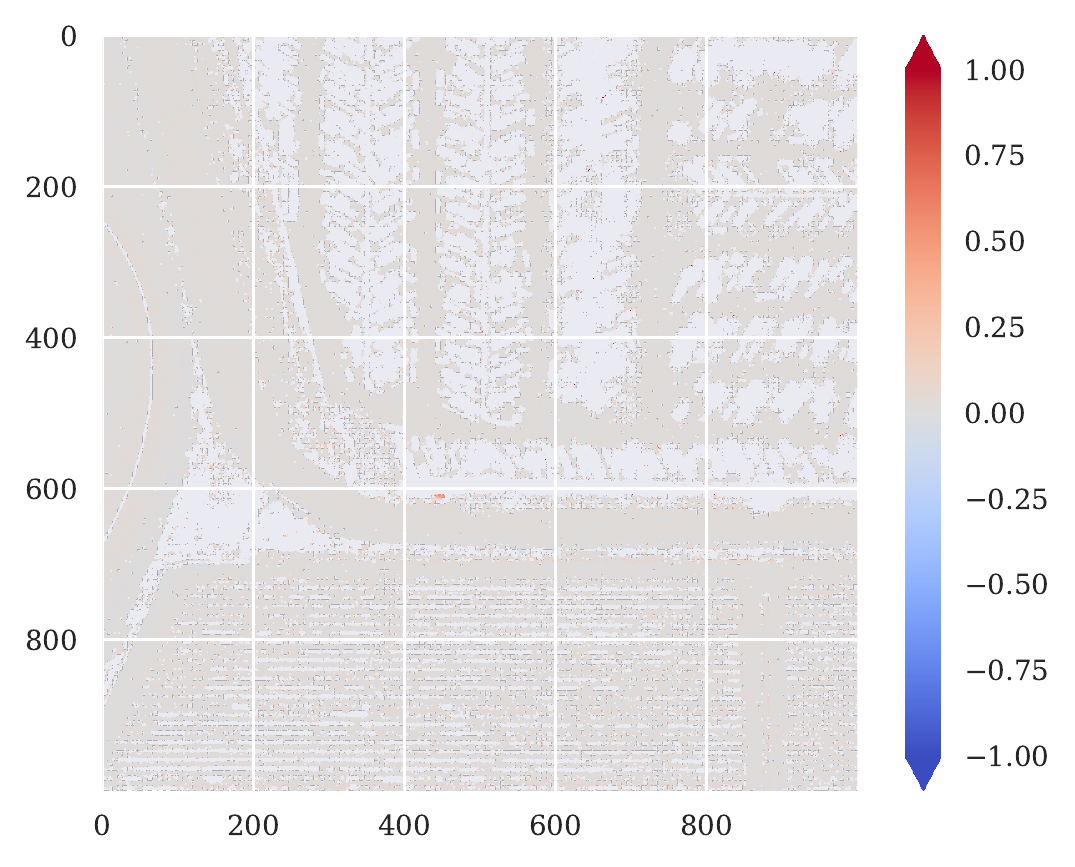}}%
        \caption{
        DALES dataset alignment: visualization of ground elevation difference (in meters) between \textit{voxel-bottom} raster and provided DTM without (a) and with (c) offset correction; and (b) visualization of RMSEs for different offsets during a grid search with minimum value at $(3, 5)$. }
    \label{fig:offset}
\end{figure*}

\begin{figure}
    \centering
    \includegraphics[width=0.7\columnwidth]{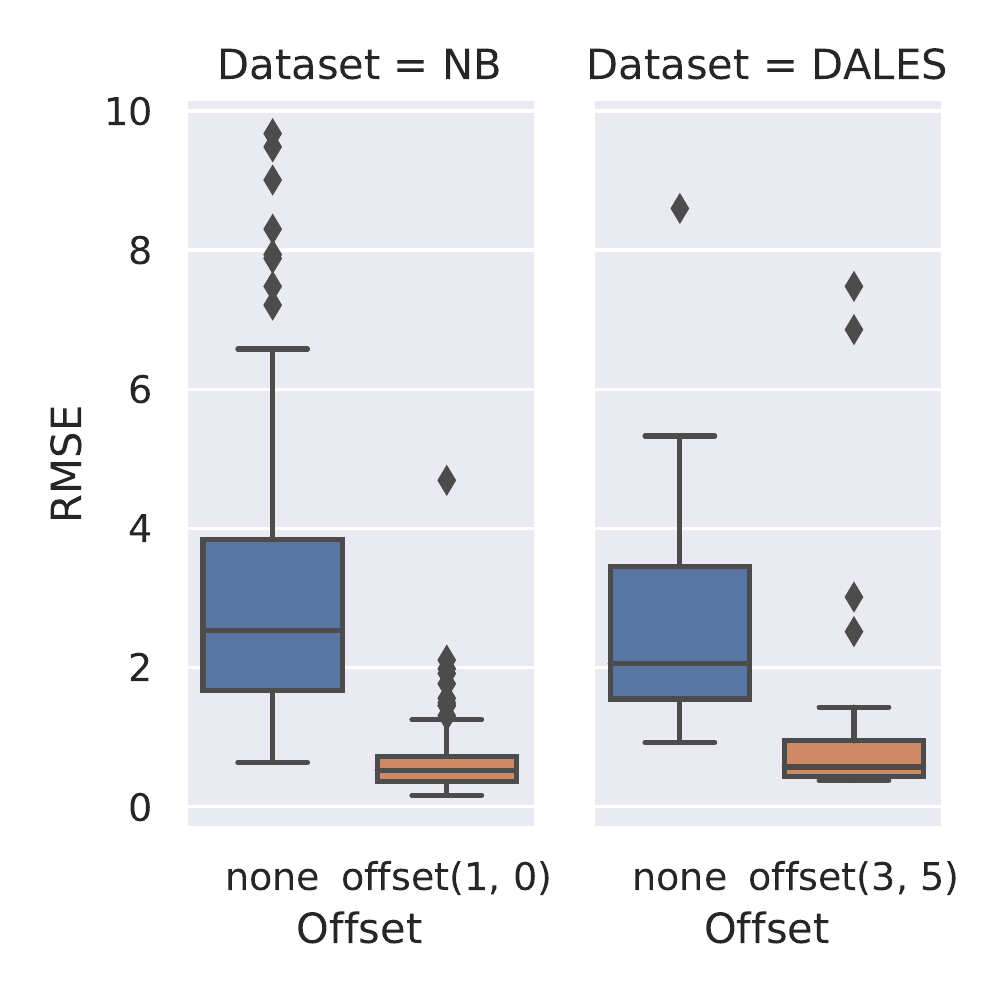}
    \caption{Per-tile alignment evaluation: median RMSEs of NB and DALES 
    subset, respectively, decrease from 2.53, 2.13 (before correction, blue 
    boxes) to 0.52, 0.59  (after correction, orange boxes), respectively. }
    \label{fig:offset_box}
\end{figure}

\subsection{GAN-based DTM generation}
\label{subsec:deepterra}

The DTM generation is formulated as an image-to-image translation problem~\cite{Isola2017}.
The pairs $(r,t)$ of generated rasters and corresponding reference DTMs are used to train a 
generative adversarial network (GAN).
As the paper does not contribute to the GAN architecture but rather on the usage of rasterization for DTM extraction, we briefly describe the architecture and refer interested readers to the original work for more details. 

The network includes a generator $\mathcal{G}$
which tries to produce, from the input rasters $r$, a DTM $\mathcal{G}(r)$ as
close as possible to the reference one $t$, and a discriminator $\mathcal{D}$,
which is trained to distinguish a generated DTM $\mathcal{G}(r)$ from the real one $t$.
The detailed architecture of each network is shown in Figure~\ref{fig:pipeline}.

\mysubsection{The generator } is taken after the off-the-shelf 
U-Net architecture~\cite{unet}, typically for pixel-to-pixel problems.
\review{
The input rasters are fed through series of several convolutional layers
and non-linearity which reduce the spatial dimensions to learn the raster
features before upsampling to the same dimensions (see Fig~\ref{fig:pipeline}).
The generator outputs an estimated DTM of the same width and height as the input
rasters.
}

The output DTMs are compared to the reference DTMs using a L1-loss given by
\begin{equation}
    \mathcal{L}_{\mathcal{G}} = \left\|t - \mathcal{G}(r)\right\|_1
\end{equation}

\review{
Euclidean losses such as L1 or L2 are well known for only being able to capture
low-frequency information and producing blurry
images~\cite{Larsen2016,Isola2017}. On the other hand, the pixel-wise
comparison does not guarantee similar statistics of the generated image
to the ground truth, hence might distorts the ground surfaces.
Thus an adversarial loss is proposed.
}

\mysubsection{The discriminator} comprises a simple architecture adapted
from~\cite{Radford2016} with conditional GAN idea from~\cite{Isola2017} and
patchGAN from~\cite{Li2016} which has proven to better capture local statistics.
\review{
In a conditional GAN, the discriminator receives a DTM, generated or
ground truth, together with the input rasters and learns to identify the fake from
the real DTMs. The extra rasters input allow the discriminator to condition the
output based on each particular input and hence improves its judging ability.
The patchGAN loss penalizes the discriminator output at the scales of image patches,
thus enforcing local statistical correctness.
}
The adversarial loss is computed from the discriminator output as follows
\begin{equation}
    \label{eq:lsgan}
    \mathcal{L}_{\mathcal{D}} = \log\mathcal{D}(r, t) + \log(1 - \mathcal{D}(r, \mathcal{G}(t))).
\end{equation}
The final loss is 
\begin{equation}
    \label{eq:loss}
    \mathcal{L} = \lambda_\mathcal{G}\mathcal{L}_\mathcal{G} + \lambda_\mathcal{D}\mathcal{L}_\mathcal{D},
\end{equation}

where $\lambda_\mathcal{C}$ and $\lambda_\mathcal{D}$ are particular weights
for each loss, and respectively set to 100 and 1 in~\cite{Isola2017}.
The optimization is alternated between $\mathcal{G}$ and $\mathcal{D}$.

\section{Dataset}

\begin{figure}[t]
    \centering
    \includegraphics[width=0.85\linewidth]{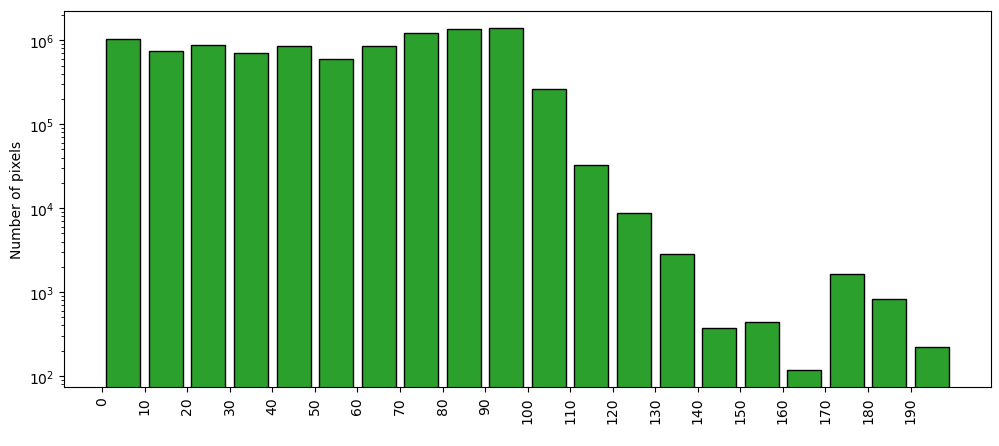} \\
    \includegraphics[width=0.85\linewidth]{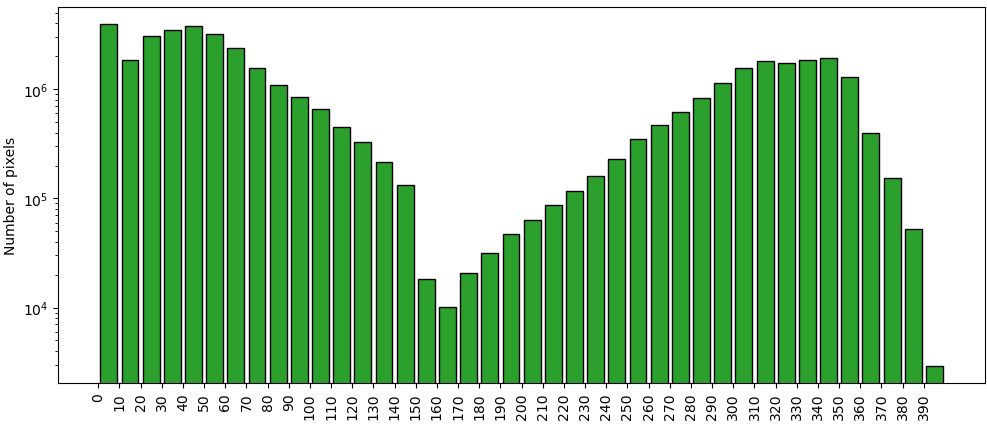}
    \caption{Elevation histogram for DALES (\emph{top}) and NB (\emph{bottom})
    subset: the x-axis depicts elevation, and y-axis the number of pixels. NB has
    a larger elevation range than DALES due to forested and mountainous regions while maintaining
    the similar shape of lower elevations in urban areas.}
    \label{fig:hist}
\end{figure}

\begin{figure}[t]
    \centering
    \includegraphics[width=0.85\linewidth]{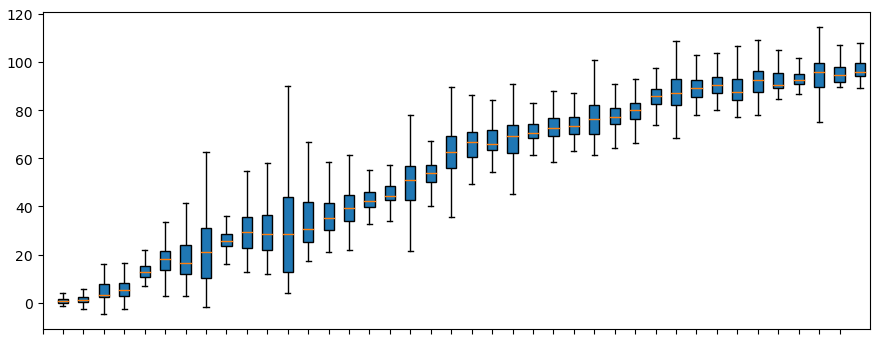} \\
    \includegraphics[width=0.85\linewidth]{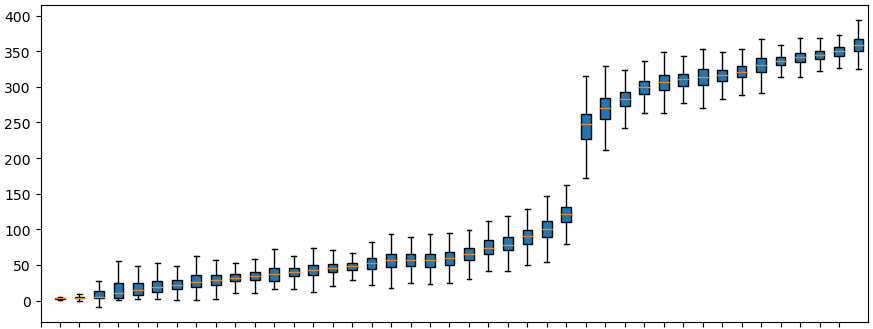}
    \caption{Elevation distribution among the tiles of DALES (\emph{top}) and NB (\emph{bottom}) subset:
    the x-axis depicts the tiles, and y-axis elevation.
    There are more variation within each tile of DALES while NB contains higher variation among the tiles.}
    \label{fig:bplot}
\end{figure}

To accommodate CNN training for DTM generation on ALS point clouds, we collect from 
open sources a large-scale dataset of ALS point clouds with reference
DTM correspondences. The dataset contains 2 subsets, DALES and New Brunswick.
Table~\ref{tab:overview} provides an overview of the data while 
general and per-tile elevation statistics of each subset are shown in Figure~\ref{fig:hist}
and Figure~\ref{fig:bplot}, respectively.

\mysubsection{DALES.} The Dayton Annotated LiDAR Earth Scan (DALES) dataset was presented
to assist on-going research for ALS point clouds~\cite{dales2020}. Here we augment it with reference DTM data, collected from the original data source of the City of
Surrey\footnote{
\url{https://data.surrey.ca/dataset/elevation-grid-2018}, 
Open Government License, City of Surrey
}.

\mysubsection{NB.}
The second subset is collected from the New Brunswick (NB)
region, open data collection, which includes
LiDAR\footnote{
\url{http://geonb.snb.ca/li/},
Open Government License, New Brunswick
} and corresponding
DTM\footnote{
\url{http://geonb.snb.ca/nbdem/}, 
Open Government License, New Brunswick, 
}.
To vary the elevation, we sample the data tiles
around the Saint John regions (urban and rural areas), and the Fundy National Park
(forested and mountainous area).

\begin{table}[t]
    \centering
    \setlength{\tabcolsep}{2pt}
    \begin{tabular}{@{}lcccccc@{}}
        \toprule
          &  coverage & \#points & tile size & DTM res.  & density & semantics \\
        \midrule
        DALES & 10 km$^2$ &  505M  & $(500$ m$)^2$ & $(25$ cm$)^2$ & 50.5 ppm${^2}$ & $\checkmark$ \\
        NB    & 42 km$^2$ & 1153M  & $(1$ km$)^2$ & $(1$ m$)^2$ & 27.4 ppm${^2}$ & $\checkmark$ \\
        \bottomrule
    \end{tabular}
    \caption{Overview of ALS point clouds in DALES and NB subset. The point clouds are
    also accompanied by standard laser-scanning fields such as point-source id,
    echo numbers, scan angle rank,~\etc}
    \label{tab:overview}
\end{table}

\review{
The ALS point clouds are produced using the latest boresight values from
the acquiring sensors (Terrascan for DALES) with preliminary quality
assurance steps taken to ensure data integrity.
Classification is carried out by combination of automatic and manual processes
followed by manual and visual QC pass inspection. Several iterations are performed
for fine-tuning and separating particular classes.
The final bare earth DTMs are produced by creating 
triangular irregular networks (TIN) on the ground and keypoint classes.
}

A quality-control step is performed to enforce the alignment between the 
point cloud rasters and provided DTMs in each subset.
Due to possible numerical instability or different rasterizing coordinations,
there are small offsets between a point cloud raster and its corresponding DTM.
To quantify, the root-mean-square error (RMSE) is computed between a raster and
provided DTM on the ground regions using ground truth semantics.
Figure~\ref{fig:offset:no} showcases an example, while 
Figure~\ref{fig:offset_box} (blue boxes) shows the distribution of each subset
(median values of 2.13 for NB and 2.53 for DALES).
To correct the misalignment, a brute-force grid search is proposed. 
We search for both vertical and horizontal correction in the integer range of $[0, 10]$ and pick the pairs that minimize the errors.

The error curves for DALES are graphed in
Figure~\ref{fig:offset:rmse}. The local minimums for NB and DALES are at $(1, 0)$ and $(3, 5)$ with median RMSE of $0.52$ and $0.59$, respectively.
Figure~\ref{fig:offset_box} shows the error statistics before and after correction.
The corrected DALES example is shown in Figure~\ref{fig:offset:yes}.

\section{Experiments}
\label{sec:experiments}

\subsection{Setup}
\label{subsec:setup}

For analysis purposes, unless stated otherwise, the validation split of each subset
is used. Each experiment is performed three times and the average of the best results are
reported. The root-mean-square error (RMSE, in meters) is used for evaluation:
lower is better.
For reporting on the test splits, the single best iteration (among the 3 runs)
from the validation split is used to show generalizability and avoid overfitting.

\review{
The network is trained, validated, and tested separately on the DALES and New Brunswick (NB)
datasets. As shown in Table~\ref{tab:overview}, to mitigate the coverage difference
of DALES and NB, the original NB point clouds are split into 4 quarters, of which
2 are chosen randomly for training, 1 for validation, and 1 for testing, resulting in 84, 42,
42 input tiles of 500m squared.
The DALES subset contains 29 500-m tiles for training and 11 for testing, provided by the
authors~\cite{dales2020}.
The voxel size for both sets is set to 25-cm cubed, resulting in 2000-pixel rasters which are
downsampled 1:4 before inputting to the network.

In all the experiments, unless stated otherwise, the hyper-parameters are set following
the default set up from~\cite{Isola2017},~\ie $\lambda_\mathcal{C}=100,\lambda_\mathcal{D}=1$,
Adam optimizer with momentum parameters $\beta_1 = 0.5, \beta_2 = 0.999$ and
learning rate of $0.0002$. The optimization is alternated between $\mathcal{G}$ and $\mathcal{D}$.
}
\rebuttal{The networks are trained on a cluster node with
    8-core Intel Xeon E5-2620 @ 2.10 GHz, 64GB RAM, and RTX-2080 Ti / 11 GB VRAM.
}

\subsection{Elevation rasters as DTM}

The pixel-mean, voxel-top, and voxel-bottom rasters particularly provide
elevation information, hence can be used directly as prediction for DTMs. 
Knowing their performance gives more insight into the rasterization
strategies and the network processing them. 
In this experiment, we show their performance on both validation and
test splits. The results are given in Table~\ref{tab:prelim}.

\begin{table}[t]
    \centering
    \begin{tabular}{@{}lcccc@{}}
        \toprule
          &  \multicolumn{2}{c}{DALES} & \multicolumn{2}{c}{NB} \\
          \cmidrule(r){2-3}\cmidrule(l){4-5}
          &  val & test & val & test \\
        \midrule
        pixel-mean   &     4.40 &     4.90 &     4.92 &     4.75 \\
        voxel-top    &     5.80 &     6.31 &     7.00 &     6.76 \\
        voxel-bottom & \bf 3.16 & \bf 3.68 & \bf 2.03 & \bf 2.01 \\
        \bottomrule
    \end{tabular}
    \caption{RMSE (in meters) for using elevation rasters directly as predicted DTM
    (no other operations are performed) on validation and test split.}
    \label{tab:prelim}
\end{table}

\begin{table}[t]
    \centering
    \setlength{\tabcolsep}{2pt}
    \begin{tabular}{@{}llccccc@{}}
        \toprule
         \phantom{C} & &  - & +density & +echoes & +stdev & +3stat \\
        \midrule
        \multicolumn{6}{l}{\emph{DALES}} \\
        & pixel-mean  &     1.75 &     1.60 &     1.54 &     1.42 &     1.24 \\
        & voxel-top   &     1.64 &     1.61 &     1.62 &     1.54 & \bf 1.12 \\
        & voxel-bottom&     1.72 &     1.60 &     1.56 &     1.63 &     1.27 \\
        & 3elev       & \bf 1.39 & \bf 1.31 & \bf 1.27 & \bf 1.45 &     1.23 \\
        \midrule
        \multicolumn{6}{l}{\emph{NB}} \\
        & pixel-mean   &     2.30 &     2.43 &     2.01 & \bf 1.83 &     1.46 \\
        & voxel-top    &     2.70 &     2.82 &     2.18 &     1.85 &     1.29 \\
        & voxel-bottom & \bf 1.97 & \bf 1.81 & \bf 1.55 &     1.85 & \bf 1.22 \\
        & 3elev        &     2.13 &     2.20 &     1.86 &     2.06 &     1.71 \\
        \bottomrule
    \end{tabular}
    \caption{\rebuttal{RMSE (in meters) of predicted DTMs from rasters combination on the
    DALES and NB validation set:
    the elevation rasters (rows, last for all) are concatenated with statistic
    rasters (columns, first for none, last for all). Adding statistic rasters
    improves the network performance. }}
    \label{tab:val_DALES_NB}
\end{table}

\begin{table}[t]
    \centering
    \setlength{\tabcolsep}{2pt}
    \begin{tabular}{@{}llcccccc@{}}
        \toprule
          \phantom{C}& &  \multirow{2}{*}{-} & \multirow{2}{*}{+sem1} & \multirow{2}{*}{+sem2} & \multicolumn{3}{c}{+3stat} \\
        \cmidrule(l){6-8}
          & & &  &  & - & +sem1 & +sem2 \\
        \midrule
        \multicolumn{6}{l}{\emph{DALES}} \\
            & voxel-top    & \bf 1.64 &    1.59 &     1.20 & \bf 1.12 &     1.06 & \bf 0.88 \\
            & voxel-bottom & 1.72 & \bf 1.17 & \bf 0.88 &     1.27 & \bf 0.97 &     0.91 \\
        \midrule
        \multicolumn{6}{l}{\emph{NB}} \\
            & voxel-top    & 2.70 &     2.03 &     2.02 &     1.29 &     1.25 &     1.22 \\
            & voxel-bottom & \bf 1.97 & \bf 1.76 & \bf 1.19 & \bf 1.22 & \bf 1.19 & \bf 1.06 \\
        \bottomrule
    \end{tabular}
    \caption{\rebuttal{RMSE (in meters) of predicted DTMs from rasters and semantic combination on the
    DALES and NB validation set: the elevations (rows) are concatenated with 1-channel (sem1) or
    2-channel (sem2) \emph{binary}. Voxel-bottom combined with 2-channel semantic and 
    statistic rasters shows superior performance.}}
    \label{tab:sem_DALES_NB}
\end{table}

\begin{table}[t]
    \centering
    \begin{tabular}{@{}llcc@{}}
        \toprule
          &&  DALES & NB \\
        \midrule
            \multicolumn{4}{l}{\textit{Rule-based}} \\
            & PMF~\cite{Zhang2003}   &     4.27 &      0.81 \\
            & SBM~\cite{Bartels2010} &     3.59 &      4.15 \\
            & SMRF~\cite{Pingel2013} &     4.08 & \bf  0.75 \\
            & CSF~\cite{Zhang2016}   & \bf 1.19 &      0.88 \\
        \midrule
            \multicolumn{4}{l}{\textit{Deep-learning-based}} \\
            &Elevation rasters           &      1.98 &      2.05 \\
            &Elevation+3stat             &      1.27 &      1.31 \\
            &Elevation+3stat+sem2        & \bf  0.82 & \bf  0.98 \\ %
        \bottomrule
    \end{tabular}
    \caption{RMSE (in meters) of the proposed deep-learning-based method in comparison with existing rule-based approaches on DALES and NB test sets.}
    \label{tab:sota}
\end{table}

\begin{figure*}[t]
    \centering
    \begin{tikzpicture}
        \node at (0, 0) {
            \includegraphics[width=\linewidth]{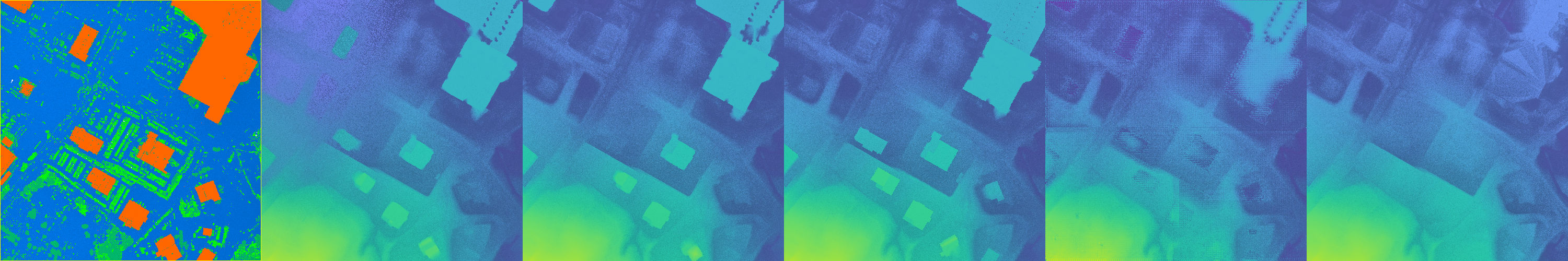}
        };
        \node at (0, -3.0) {
            \includegraphics[width=\linewidth]{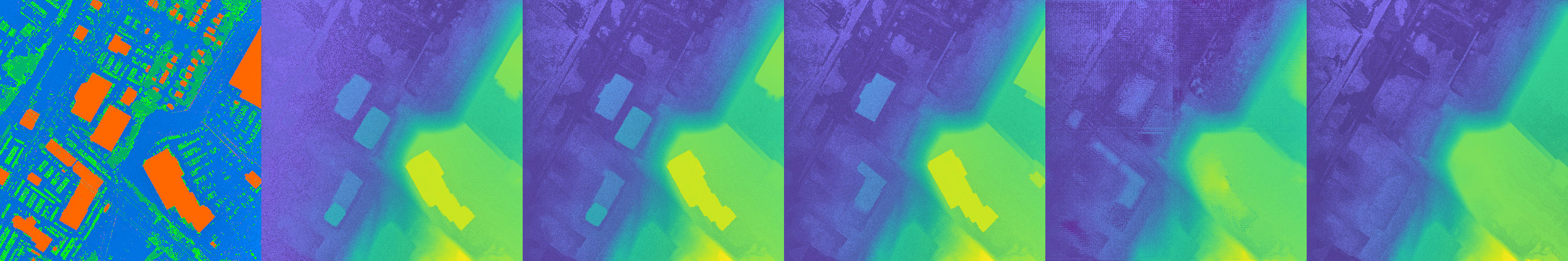}
        };
        \node at (0, -6.0) {
            \includegraphics[width=\linewidth]{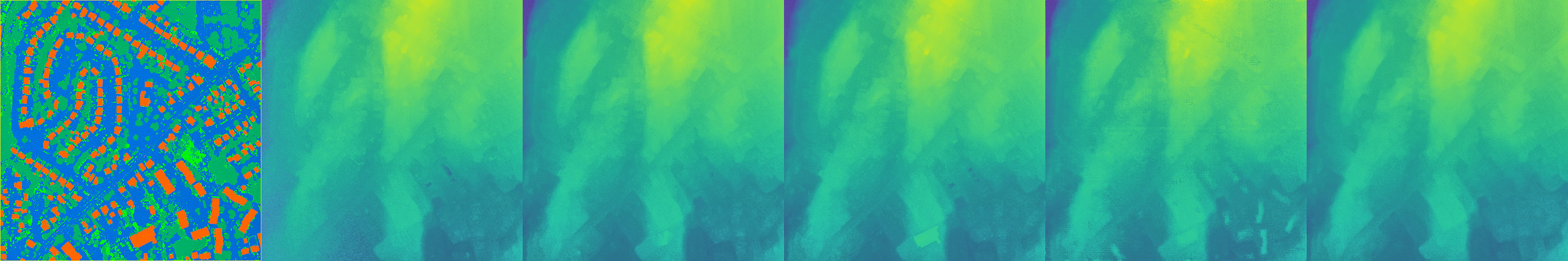}
        };
        \node at (-7.2,1.7) {\small Point cloud (top)};
        \node at (-4.3,1.7) {\small PMF};
        \node at (-1.4,1.7) {\small SMRF};
        \node at ( 1.4,1.7) {\small CSF};
        \node at ( 4.3,1.7) {\small Elev+Stats};
        \node at ( 7.2,1.7) {\small Ground truth};
    \end{tikzpicture}
    \caption{RMSE (in meters) on the test split: the deep-learning-based method
    with elevation and statistics rasters (Elev+Stats) removes most of the objects
    and shows better performance on large-scale buildings.
    The point clouds are shown with semantic labels: ground (blue), buildings 
    (red), vegetation (green).
    }
    \label{fig:qualitative}
\end{figure*}
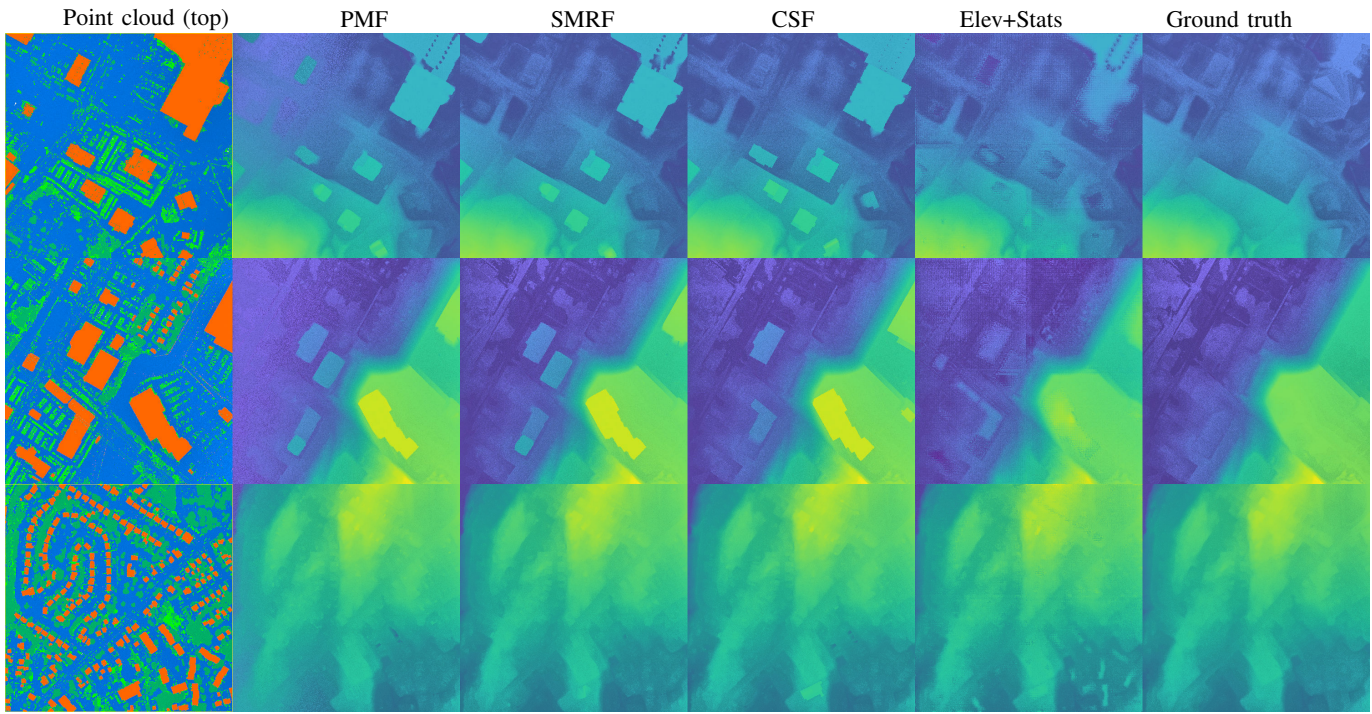

From the definitions, it is trivial that voxel-bottom rasters are the
closest to DTMs, while voxel-top produces the highest errors, and pixel-mean stays
in between. 
The rasters errors also provide insights of the object types
and heights in each set, hinting the object semantics.
As laser beams can penetrate well in vegetation
areas but are blocked by solid objects,~\eg buildings, voxel-bottom
errors depict the presence of buildings and large constructions. The results show that
there are more and/or higher buildings in the DALES subset than in NB. On the
other hand, voxel-top rasters pick the highest object points, thus show that
NB might comprise more, and/or higher, trees. They agree with the fact
that the DALES subset is focused on urban areas and NB contains more forested regions.
Such information may be used to quantitatively identify scene category without
the availability of semantic labels.

\subsection{Combination of rasters}
\label{subsec:exp1}

In this experiment, the elevation rasters (pixel-mean, voxel-top, and voxel-bottom)
are employed for DTM extraction. Several strategies are possible
including (1) using each elevation raster individually, (2) in combination,
together with (3) one or (4) all statistic rasters (density, stdev, and echoes). 
The results for different combination strategies are shown in Table~\ref{tab:val_DALES_NB}.

\review{
In general, adding all 3 statistics rasters improves the network performance
by a large margin. Hence this could be the go-to solution. Regarding the
elevation rasters to be used, the current results show the dependency to the
type of scenes: as NB contains a dominantly large number of vegetation
(mountainous and forested regions), the voxel-bottom rasters provide
a close representation to DTMs while it is closer to voxel-top rasters for
large constructions in urban areas from the DALES subset.
}
Thus, the combination of all 3 elevation rasters dominates in DALES while
voxel-bottom dominates in NB. The performance gap is diminished when adding
statistics information. For later experiments, only voxel-top and voxel-bottom
are used with statistics rasters.

\subsection{Loss coefficients}
\label{subsec:ablation}

\review{
In this experiment, we show how the losses contribute to the network performance.
We vary the GAN loss coefficient $\lambda_\mathcal{D}$ with respect to 2 values
of the L1 loss coefficient $\lambda_\mathcal{G}$.
As shown in Table~\ref{tab:loss_weights}, although the 2 coefficients are not completely linearly correlated, the network performs poorly when there is no GAN loss contribution  $\lambda_\mathcal{D}=0$, showing its necessity. 
}
\begin{table}[t]
    \centering
    \begin{tabular}{@{}rr@{}}
        \toprule
        $\lambda_\mathcal{D}$  & $\lambda_\mathcal{G}=100$ \\ %
        \midrule
        0 & 2.98 \\
        0.01 & 2.68 \\
        0.1 & 2.57 \\
        1 & \bf 1.72 \\
        10 & 1.93 \\
        100 & 2.48 \\
        \bottomrule
    \end{tabular}
    \caption{\review{RMSE (in meters) for various GAN coefficients $\lambda_\mathcal{D}$ with respect to 2 values of L1 coefficients $\lambda_\mathcal{G}$ on the DALES validation set: the network performs best at $\left(\lambda_\mathcal{G}, \lambda_\mathcal{D}\right) = (100, 1)$. The networks perform poorly when there is no GAN loss contribution,~\ie $\lambda_\mathcal{D}=0$.
    }}
    \label{tab:loss_weights}
\end{table}

\subsection{Semantics}
\label{subsec:exp_semantic}

DTM extraction methods generally rely on ground filtering labels.
In this experiment, we explore the effects of semantic information
to the pipeline. 
To maintain consistency with the previous setups, we
concatenate semantic maps to the rasters as extra channels and leave
the exhaustive study of efficient usage of semantics for future work.
\review{
This is a less direct use of semantic information compared to the more
popular way in which objects' regions are removed and
inpainted~\cite{Ayhan2020,Kwan20,Crema2020}.
}
The semantic maps are generated by rasterizing the mode of point labels
in each vertical column,~\ie each pixel receives the most frequent label
in the vertical column.

The label raster is then binarized as ground/non-ground and converted to one-hot
vector format. Although a single-channel binary image (sem1) is, in principle,
sufficient for describing binary classes, we also include the 
complementary channel, making 2-channel semantic maps (sem2), for the sake of comparison
The results are shown
in Table~\ref{tab:sem_DALES_NB}.

Adding semantic labels, in general, improves the network performance. Surprisingly,
despite the redundant information, using 2-channel complementary binary labels (+sem2)
show better performance than when using 1-channel labels,
hinting how semantic maps should be handled for DTM extraction.
\review{
This could be explained that the complementary information might not be trivial
to be derived with the current network capacity and show potential improvement
when soft-labeling vectors are used,~\ie when classes are represented by floating point probabilities instead of binary numbers.
}

\subsection{Comparison with state-of-the-art}
\label{subsec:sota}

To evaluate the data-driven approach, we compare and show the \emph{test} results of our
best-validated iteration and those of the well-established ground-filtering
methods designed for DTM extraction. Specifically,
four baselines are used with default parameters, including
PMF~\cite{Zhang2003}
(1m cell, exponential window of maximum 33m, distance of 15cm-250cm, and 
slope of 45°),
SMRF~\cite{Pingel2013}
(1m cell, max window of 18m, elevation scalar of 1.25m, threshold of 0.5m,
and 15\% slope tolerance),
SBM~\cite{Bartels2010}, and
CSF~\cite{Zhang2003}
(cloth resolution of 0.5m, rigidness of 3, time step of 0.65, 500 iterations, and slope smoothing post-processing).
The implementations of PMF, SMRF and SBM are from the Point Data Abstraction
Library (PDAL)\footnote{\url{https://pdal.io/}} while CSF is provided by
the original authors\footnote{\url{https://github.com/jianboqi/CSF}}. 
A triangular irregular network (TIN) is computed using Delaunay triangulation
on the extracted ground points before being rasterized to DTMs.
The results of rule-based methods provided in Table~\ref{tab:sota} are also meant for
benchmarking our dataset.

As a parameter-free method, SBM relies on strong assumptions of flat ground
and sacrifices its performance. PMF, SMRF, and CSF perform similarly well on the
NB subset with scattered small houses and forested regions. While PMF and SMRF
suffer for large buildings and constructions on the DALES subset, CSF shows
more robust results even with fixed default parameters.

For comparison with deep-learning-based methods, we use the best iteration on
validation set from previous experiments.
The deep-learning-based methods show more stable  
similar results on both DALES and NB subset. Due to independence from fixed
parameters, deep-learning-based methods rely only on training set variations,
thus can, in principle, be improved when more data are collected.

The image-to-image approach, however, is limited due to the loss of point
cloud geometry of which the rule-based methods are designed to take advantage.
Yet as more information is provided by the statistical rasters,
the agnostic off-the-shelf network's performance approaches that of the best rule-based
designed directly for ground-filtering.

Qualitative results in Figure~\ref{fig:qualitative} show that a network with inputs of
elevation and statistical rasters is more robust to large objects. Most of the houses
and large buildings that left untouched by the rule-based methods on the first 2 rows are
removed by the network.
As the rule-based methods depend greatly of fixed parameters, they might be effective
with objects of certain sizes (small objects on the third row),
but failed in other cases: it is well-known that the parameters need to be carefully
tuned for separate cases, showing needs for a deep-learning-based and end-to-end solution.

\section{Conclusion}
\label{sec:conclusion}

This paper attempts to unify the DTM extraction process from ALS point clouds using a
deep-learning-based method. By formulating the problem as an image-to-image translation
and using rasterization techniques, off-the-shelf computer-vision-designed architectures can be used
to tackle this important remote sensing task.
To stimulate data-driven approaches on this problem,
a large-scale dataset of ALS point clouds and reference DTMs is also introduced and evaluated.

The proposed method, despite being domain-agnostic, shows comparable results to task-dedicated methods and performs on par with state-of-the-art baselines.
Experiments show that the various point-cloud-specific qualities can be gathered by simple rasterization
techniques, which have proven to be helpful for predicting bare-ground elevation.

Among all types of information to be rasterized, semantic play an important role in extracting DTM due to its
ability in identifying objects. Although in the paper, only one-hot vectors ground truth labels are used, the
same idea could be applied to softmax-smooth predicted labels. Other effective ways of exploiting semantic
information for DTM extract are encouraged for future research.

Effective may it be, rasterization nonetheless squashes geometry-rich 3D point clouds into 2D representations, hence can only depict a certain aspects of a point cloud. As such, this aggressive operation,
together with the 2D image-to-image translation approach, does not fully exploit the potential of 3D data. The sub-metric performance gap with other ground filters which rely on point cloud geometry show
the room for development of deep-learning-based method for DTM extraction in future research.

\section*{Acknowledgment}

This work was supported by the ANR SixP project under the reference ANR-19-CE02-0013.

\ifCLASSOPTIONcaptionsoff
  \newpage
\fi

\bibliographystyle{IEEEtran}
\bibliography{IRISA-EV21.bib}

\begin{thebibliography}{10}
\providecommand{\url}[1]{#1}
\csname url@samestyle\endcsname
\providecommand{\newblock}{\relax}
\providecommand{\bibinfo}[2]{#2}
\providecommand{\BIBentrySTDinterwordspacing}{\spaceskip=0pt\relax}
\providecommand{\BIBentryALTinterwordstretchfactor}{4}
\providecommand{\BIBentryALTinterwordspacing}{\spaceskip=\fontdimen2\font plus
\BIBentryALTinterwordstretchfactor\fontdimen3\font minus
  \fontdimen4\font\relax}
\providecommand{\BIBforeignlanguage}[2]{{%
\expandafter\ifx\csname l@#1\endcsname\relax
\typeout{** WARNING: IEEEtran.bst: No hyphenation pattern has been}%
\typeout{** loaded for the language `#1'. Using the pattern for}%
\typeout{** the default language instead.}%
\else
\language=\csname l@#1\endcsname
\fi
#2}}
\providecommand{\BIBdecl}{\relax}
\BIBdecl

\bibitem{Bakula2011}
K.~Baku{\l}a, ``{Reduction of DTM obtained from LiDAR data for flood
  modeling},'' \emph{Archives of Photogrammetry}, vol.~22, pp. 51--61, 2011.

\bibitem{Muhadi2020}
N.~A. Muhadi, A.~F. Abdullah, S.~K. Bejo, M.~R. Mahadi, and A.~Mijic, ``{The
  Use of LiDAR-Derived DEM in Flood Applications: A Review},'' \emph{Remote
  Sensing}, vol.~12, no.~14, 2020.

\bibitem{Muthusamy2021}
M.~Muthusamy, M.~{Rivas Casado}, D.~Butler, and P.~Leinster, ``{Understanding
  the effects of Digital Elevation Model resolution in urban fluvial flood
  modelling},'' \emph{Journal of Hydrology}, p. 126088, 2021.

\bibitem{Gonzalez2019}
V.~Gonz{\'{a}}lez-Jaramillo, A.~Fries, and J.~Bendix, ``{AGB Estimation in a
  Tropical Mountain Forest (TMF) by Means of RGB and Multispectral Images Using
  an Unmanned Aerial Vehicle (UAV)},'' \emph{Remote Sensing}, vol.~11, no.~12,
  2019.

\bibitem{Xu2021}
Z.~Xu, Z.~Shen, Y.~Li, L.~Xia, H.~Wang, S.~Li, S.~Jiao, and Y.~Lei, ``{Road
  Extraction in Mountainous Regions from High-Resolution Images Based on DSDNet
  and Terrain Optimization},'' \emph{Remote Sensing}, vol.~13, no.~1, 2021.

\bibitem{Gevaert2018}
C.~M. Gevaert, C.~Persello, F.~Nex, and G.~Vosselman, ``{A deep learning
  approach to DTM extraction from imagery using rule-based training labels},''
  \emph{ISPRS Journal of Photogrammetry and Remote Sensing}, vol. 142, pp.
  106--123, 2018.

\bibitem{Rokhmana2019}
C.~A. Rokhmana and A.~R. Sastra, ``{DTM generation from aerial photo and
  Worldview-3 images by using different DSM filtering methods},'' in
  \emph{Sixth Geoinformation Science Symposium}, vol. 11311.\hskip 1em plus
  0.5em minus 0.4em\relax SPIE, 2019, pp. 152--158.

\bibitem{Duan2019}
L.~Duan, M.~Desbrun, A.~Giraud, F.~Trastour, and L.~Laurore, ``{Large-Scale DTM
  Generation From Satellite Data},'' in \emph{Proceedings of the {IEEE}/{CVF}
  Conference on Computer Vision and Pattern Recognition Workshop (CVPRw)},
  2019.

\bibitem{Debella2016}
M.~Debella-Gilo, ``{Bare-earth extraction and DTM generation from
  photogrammetric point clouds including the use of an existing
  lower-resolution DTM},'' \emph{International Journal of Remote Sensing
  (IJRS)}, vol.~37, no.~13, pp. 3104--3124, 2016.

\bibitem{Ye2019}
W.~Ye, ``{Extraction of Digital Terrain Models from Airborne Laser Scanning
  Data based on Transfer-Learning},'' Master's thesis, University of Waterloo,
  2019.

\bibitem{dublin2019}
I.~Zolanvari, S.~Ruano, A.~Rana, A.~Cummins, R.~E. da~Silva, M.~Rahbar, and
  A.~Smolic, ``{DublinCity: Annotated LiDAR Point Cloud and its
  Applications},'' in \emph{Proceedings of the British Machine Vision
  Conference (BMVC)}, 2019.

\bibitem{dales2020}
N.~Varney, V.~K. Asari, and Q.~Graehling, ``{DALES: A Large-scale Aerial LiDAR
  Data Set for Semantic Segmentation},'' in \emph{Proceedings of the
  {IEEE}/{CVF} Conference on Computer Vision and Pattern Recognition Workshop
  (CVPRw)}, 2020, pp. 186--187.

\bibitem{lasdu2020}
Z.~Ye, Y.~Xu, R.~Huang, X.~Tong, X.~Li, X.~Liu, K.~Luan, L.~Hoegner, and
  U.~Stilla, ``{LASDU: A Large-scale Aerial LiDAR Dataset for Semantic Labeling
  in Dense Urban Areas},'' \emph{ISPRS International Journal of
  Geo-Information}, vol.~9, no.~7, 2020.

\bibitem{opengf2021}
N.~Qin, W.~Tan, L.~Ma, D.~Zhang, and J.~Li, ``Opengf: An ultra-large-scale
  ground filtering dataset built upon open {ALS} point clouds around the
  world,'' \emph{Proceedings of the IEEE/CVF Conference on Computer Vision and
  Pattern Recognition Workshops}, 2021.

\bibitem{Krizhevsky2012}
A.~Krizhevsky, I.~Sutskever, and G.~E. Hinton, ``{ImageNet Classification with
  Deep Convolutional Neural Networks},'' in \emph{Proceedings of the Conference
  on Neural Information Processing Systems (NeurIPS)}, USA, 2012, pp.
  1097--1105.

\bibitem{Long2015}
J.~Long, E.~Shelhamer, and T.~Darrell, ``{Fully convolutional networks for
  semantic segmentation},'' in \emph{Proceedings of the {IEEE}/{CVF} Conference
  on Computer Vision and Pattern Recognition (CVPR)}, 2015, pp. 3431--3440.

\bibitem{rcnn}
R.~Girshick, J.~Donahue, T.~Darrell, and J.~Malik, ``{Rich feature hierarchies
  for accurate object detection and semantic segmentation},'' in
  \emph{Proceedings of the {IEEE}/{CVF} Conference on Computer Vision and
  Pattern Recognition (CVPR)}, 2014, pp. 580--587.

\bibitem{Hu2016}
X.~Hu and Y.~Yuan, ``{Deep-Learning-Based Classification for DTM Extraction
  from ALS Point Cloud},'' \emph{Remote Sensing}, vol.~8, no.~9, 2016.

\bibitem{Zhang2020}
J.~Zhang, X.~Hu, H.~Dai, and S.~Qu, ``{DEM Extraction from ALS Point Clouds in
  Forest Areas via Graph Convolution Network},'' \emph{Remote Sensing},
  vol.~12, no.~1, 2020.

\bibitem{Guiotte2020}
F.~Guiotte, M.~Pham, R.~Dambreville, T.~Corpetti, and S.~Lef{\`{e}}vre,
  ``Semantic segmentation of lidar points clouds: Rasterization beyond digital
  elevation models,'' \emph{{IEEE} Geoscience Remote Sensing Letters}, vol.~17,
  no.~11, pp. 2016--2019, 2020.

\bibitem{Isola2017}
P.~Isola, J.-Y. Zhu, T.~Zhou, and A.~A. Efros, ``{Image-to-Image Translation
  with Conditional Adversarial Networks},'' in \emph{Proceedings of the
  {IEEE}/{CVF} Conference on Computer Vision and Pattern Recognition (CVPR)},
  2017, pp. 5967--5976.

\bibitem{Nguyen2018}
T.~Nguyen-Phuoc, C.~Li, S.~Balaban, and Y.-L. Yang, ``{RenderNet: A deep
  convolutional network for differentiable rendering from 3D shapes},'' in
  \emph{Proceedings of the Conference on Neural Information Processing Systems
  (NeurIPS)}, 2018.

\bibitem{GAN}
I.~Goodfellow, J.~Pouget-Abadie, M.~Mirza, B.~Xu, D.~Warde-Farley, S.~Ozair,
  A.~Courville, and Y.~Bengio, ``{Generative Adversarial Nets},'' in
  \emph{Proceedings of the Conference on Neural Information Processing Systems
  (NeurIPS)}, 2014, pp. 2672--2680.

\bibitem{Hutchinson2011}
M.~F. Hutchinson, T.~Xu, and J.~A. Stein, ``{Recent Progress in the ANUDEM
  Elevation Gridding Procedure},'' in \emph{Geomorphometry 2011}, 2011, pp.
  19--22.

\bibitem{Sithole2001}
G.~Sithole, ``{Filtering of Laser Altimetry Data using a Slope Adaptive
  Filter},'' \emph{International Archives of Photogrammetry and Remote
  Sensing}, vol.~34, no.~3, 2001.

\bibitem{Sithole2004}
G.~Sithole and G.~Vosselman, ``{Experimental comparison of filter algorithms
  for bare-Earth extraction from airborne laser scanning point clouds},''
  \emph{ISPRS Journal of Photogrammetry and Remote Sensing}, vol.~59, no.~1,
  pp. 85--101, 2004.

\bibitem{Zhang2003}
K.~Zhang, S.-C. Chen, D.~Whitman, M.-L. Shyu, J.~Yan, and C.~Zhang, ``{A
  progressive morphological filter for removing nonground measurements from
  airborne LIDAR data},'' \emph{IEEE Transactions on Geoscience and Remote
  Sensing}, vol.~41, no.~4, pp. 872--882, 2003.

\bibitem{Pingel2013}
T.~J. Pingel, K.~C. Clarke, and W.~A. McBride, ``{An improved simple
  morphological filter for the terrain classification of airborne LIDAR
  data},'' \emph{ISPRS Journal of Photogrammetry and Remote Sensing}, vol.~77,
  pp. 21--30, 2013.

\bibitem{Bartels2010}
M.~Bartels and H.~Wei, ``{Threshold-free object and ground point separation in
  LIDAR data},'' \emph{Pattern Recognition Letters}, vol.~31, no.~10, pp.
  1089--1099, 2010.

\bibitem{Zhang2016}
W.~Zhang, J.~Qi, P.~Wan, H.~Wang, D.~Xie, X.~Wang, and G.~Yan, ``{An
  Easy-to-Use Airborne LiDAR Data Filtering Method Based on Cloth
  Simulation},'' \emph{Remote Sensing}, vol.~8, no.~6, 2016.

\bibitem{Pijl2020}
A.~Pijl, J.-S. Bailly, D.~Feurer, M.~A. {El Maaoui}, M.~R. Boussema, and
  P.~Tarolli, ``{TERRA: Terrain Extraction from elevation Rasters through
  Repetitive Anisotropic filtering},'' \emph{International Journal of Applied
  Earth Observation and Geoinformation}, vol.~84, p. 101977, 2020.

\bibitem{imagenet}
J.~Deng, W.~Dong, R.~Socher, L.-J. Li, K.~Li, and L.~Fei-Fei, ``{ImageNet: A
  Large-Scale Hierarchical Image Database},'' in \emph{Proceedings of the
  {IEEE}/{CVF} Conference on Computer Vision and Pattern Recognition (CVPR)},
  2009.

\bibitem{He2015}
K.~He, X.~Zhang, S.~Ren, and J.~Sun, ``{Delving Deep into Rectifiers:
  Surpassing Human-Level Performance on ImageNet Classification},'' in
  \emph{Proceedings of the {IEEE}/{CVF} International Conference on Computer
  Vision (ICCV)}, 2015.

\bibitem{Sun2018}
D.~Sun, X.~Yang, M.-Y. Liu, J.~Kautz, and J.~K. Nvidia, ``{PWC-Net: CNNs for
  Optical Flow Using Pyramid, Warping, and Cost Volume},'' in \emph{Proceedings
  of the {IEEE}/{CVF} Conference on Computer Vision and Pattern Recognition
  (CVPR)}, 2018.

\bibitem{HALe2018}
H.-A. Le, A.~S. Baslamisli, T.~Mensink, and T.~Gevers, ``{Three for one and one
  for three: Flow, Segmentation, and Surface Normals},'' in \emph{Proceedings
  of the British Machine Vision Conference (BMVC)}, 2018.

\bibitem{Tapper2016}
G.~Tapper, ``{Extraction of DTM from Satellite Images Using Neural Networks},''
  Master's thesis, Link{\"{o}}ping University, Computer Vision, 2016.

\bibitem{resnet}
K.~He, X.~Zhang, S.~Ren, and J.~Sun, ``{Deep Residual Learning for Image
  Recognition},'' in \emph{Proceedings of the {IEEE}/{CVF} Conference on
  Computer Vision and Pattern Recognition (CVPR)}, 2016, pp. 770--778.

\bibitem{Wang2019}
Y.~Wang, Y.~Sun, Z.~Liu, S.~E. Sarma, M.~M. Bronstein, and J.~M. Solomon,
  ``{Dynamic Graph CNN for Learning on Point Clouds},'' \emph{ACM Transactions
  on Graphics (TOG)}, vol.~38, no.~5, oct 2019.

\bibitem{Ayhan2020}
B.~Ayhan, C.~Kwan, J.~Larkin, L.~Kwan, D.~Skarlatos, and M.~Vlachos,
  ``{Performance comparison of different inpainting algorithms for accurate DTM
  generation},'' in \emph{Geospatial Informatics X}, vol. 11398.\hskip 1em plus
  0.5em minus 0.4em\relax SPIE, 2020, pp. 144--159.

\bibitem{Kwan20}
C.~Kwan, D.~Gribben, B.~Ayhan, and J.~Larkin, ``{Practical Digital Terrain
  Model Extraction Using Image Inpainting Techniques},'' in \emph{Recent
  Advances in Image Restoration with Applications to Real World Problems},
  C.~Kwan, Ed.\hskip 1em plus 0.5em minus 0.4em\relax Rijeka: IntechOpen, 2020,
  ch.~8.

\bibitem{Crema2020}
S.~Crema, M.~Llena, A.~Calsamiglia, J.~Estrany, L.~Marchi, D.~Vericat, and
  M.~Cavalli, ``Can inpainting improve digital terrain analysis? comparing
  techniques for void filling, surface reconstruction and geomorphometric
  analyses,'' \emph{Earth Surface Processes and Landforms}, vol.~45, no.~3, pp.
  736--755, Mar. 2020.

\bibitem{Yu2018}
J.~Yu, Z.~Lin, J.~Yang, X.~Shen, X.~Lu, and T.~S. Huang, ``Generative image
  inpainting with contextual attention,'' in \emph{Proceedings of the IEEE
  Conference on Computer Vision and Pattern Recognition (CVPR)}, June 2018.

\bibitem{Panagiotou2020}
E.~Panagiotou, G.~Chochlakis, L.~Grammatikopoulos, and E.~Charou, ``Generating
  elevation surface from a single {RGB} remotely sensed image using deep
  learning,'' \emph{Remote Sensing}, vol.~12, no.~12, p. 2002, Jun. 2020.

\bibitem{isprs2012}
F.~Rottensteiner, G.~Sohn, J.~Jung, M.~Gerke, C.~Bailard, S.~Benitez, and
  U.~Breitkopf, ``{The ISPRS benchmark on urban object classification and 3D
  building reconstruction},'' in \emph{ISPRS Annals of the Photogrammetry,
  Remote Sensing and Spatial Information Sciences}, 2012, pp. 293--298.

\bibitem{Zhu2018}
L.~Zhu, S.~Shen, X.~Gao, and Z.~Hu, ``Large scale urban scene modeling from mvs
  meshes,'' in \emph{Proceedings of the European Conference on Computer Vision
  (ECCV)}, September 2018.

\bibitem{Han2021}
J.~Han, L.~Zhu, X.~Gao, Z.~Hu, L.~Zhou, H.~Liu, and S.~Shen, ``Urban scene lod
  vectorized modeling from photogrammetry meshes,'' \emph{IEEE Transactions on
  Image Processing}, vol.~30, pp. 7458--7471, 2021.

\bibitem{unet}
O.~Ronneberger, P.~Fischer, and T.~Brox, ``{U-Net: Convolutional Networks for
  Biomedical Image Segmentation},'' in \emph{Medical Image Computing and
  Computer-Assisted Intervention (MICCAI)}, Cham, 2015, pp. 234--241.

\bibitem{Larsen2016}
A.~B.~L. Larsen, S.~K. S\o{}nderby, H.~Larochelle, and O.~Winther,
  ``Autoencoding beyond pixels using a learned similarity metric,'' in
  \emph{Proceedings of the 33rd International Conference on International
  Conference on Machine Learning - Volume 48}, ser. ICML'16.\hskip 1em plus
  0.5em minus 0.4em\relax JMLR.org, 2016, p. 1558–1566.

\bibitem{Radford2016}
A.~Radford, L.~Metz, and S.~Chintala, ``Unsupervised representation learning
  with deep convolutional generative adversarial networks,'' in \emph{4th
  International Conference on Learning Representations (ICLR)}, 2016.

\bibitem{Li2016}
C.~Li and M.~Wand, ``{Precomputed Real-Time Texture Synthesis with Markovian
  Generative Adversarial Networks},'' in \emph{Proceedings of the European
  Conference on Computer Vision (ECCV)}, Cham, 2016, pp. 702--716.

\end{thebibliography}

\begin{IEEEbiography}[{\includegraphics[width=1in,height=1.25in,clip,keepaspectratio]{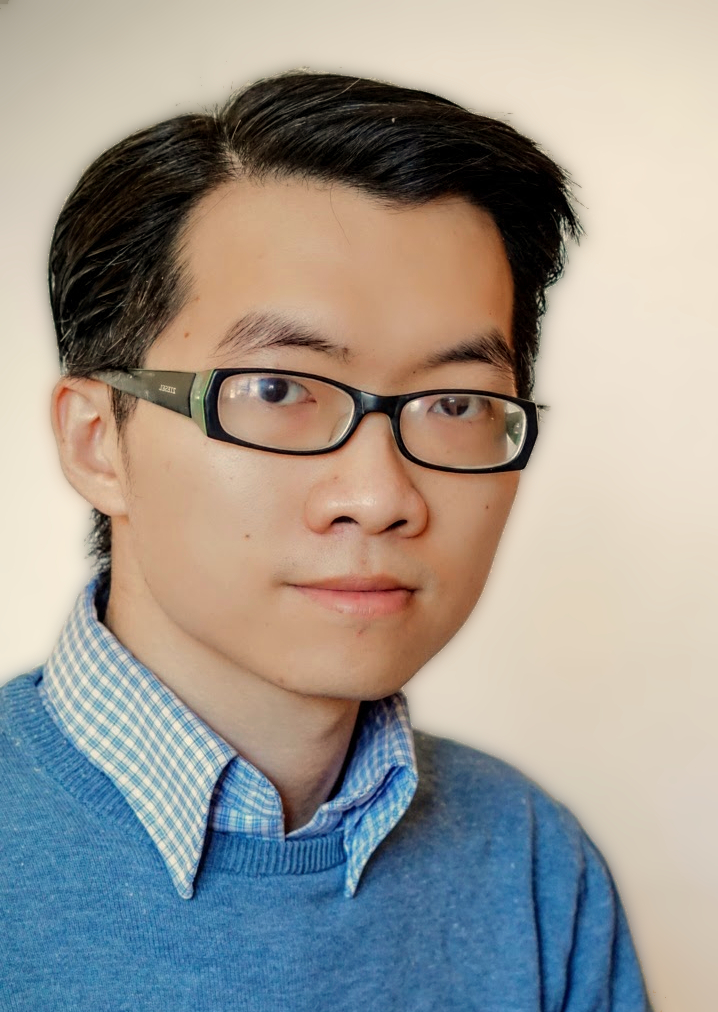}}]{Hoàng-Ân Lê}
received the B.Sc. degree in computer science in 2012 from the University of
Science, Vietnam National University, Ho Chi Minh city, Vietnam and
received the M.Sc. degree in engineering in 2015 from the EURECOM institute,
Télécom ParisTech, Paris, France. He received the Ph.D. degree in computer vision
and machine learning in 2021 from the University of Amsterdam, Amsterdam,
The Netherlands while working on the TrimBot2020 project within the EUHorizon2020 programme.
Since then, he has been a postdoctoral researcher with the OBELIX team of IRISA
Vannes at the University of South Brittany, Vannes, France.
His research interests include computer vision and machine learning, with particular
interests in multimodality-multitask learning, 3D understanding,
and applications in earth observation and remote sensing.
\end{IEEEbiography}

\begin{IEEEbiography}[{\includegraphics[width=1in,height=1.25in,clip,keepaspectratio]{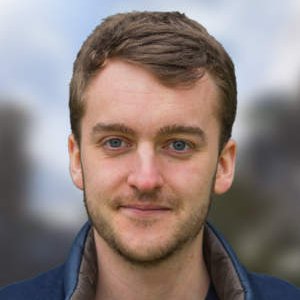}}]{Florent Guiotte}

received the M.Eng. degree in computer science in 2016 from the University Rennes~1, France.
He received the Ph.D. degree in computer science in 2021 from the University Rennes~2.
He was a post-doctoral fellow at
IRISA, OBELIX team in 2021.
Since 2022, he works for L'Avion Jaune (\href{https://lavionjaune.com}{lavionjaune.com}), Montpellier, France.
His research interests include computer vision and machine learning applied to remote sensing images and point clouds.

\end{IEEEbiography}

\begin{IEEEbiography}[{\includegraphics[height=1.25in,clip]{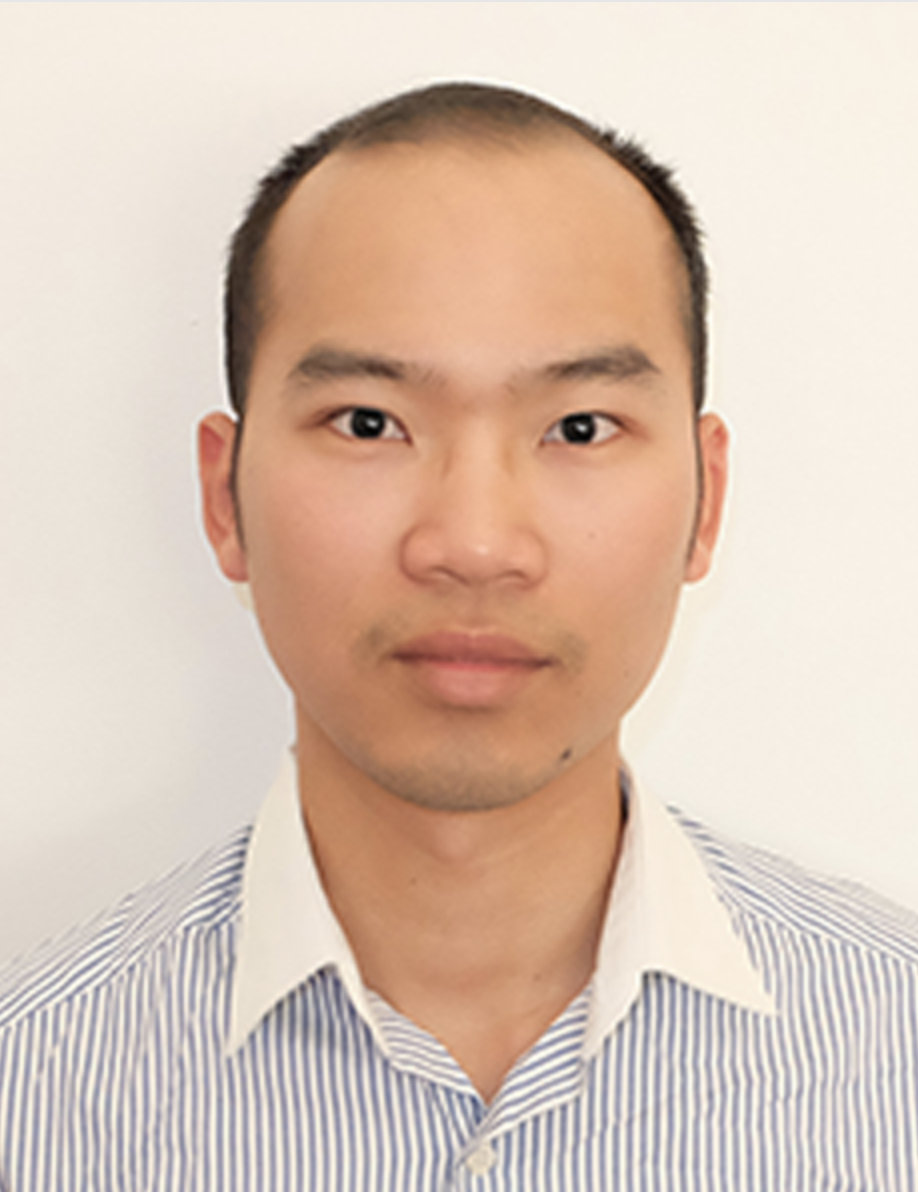}}]{Minh-Tan Pham}
obtained his M. Eng and PhD degrees from Telecom Bretagne, France in 2013 and 2016. He was an intern at the Dpt. of Geomatics, Laval University, Canada in 2013 and at the French Space Agency (CNES), France in 2015. From 2016 to 2019, he was a post-doctoral fellow at IRISA laboratory. Since 2019, he is an Assistant Professor at Univ. Bretagne Sud, IUT de Vannes and a researcher at the IRISA, OBELIX team. His research interests include image processing and machine learning applied to remote sensing data with the current focus on mathematical morphology, hierarchical representation and self-supervised deep networks for feature extraction, object detection and classification tasks.
 \end{IEEEbiography}

\begin{IEEEbiography}[{\includegraphics[width=1in,height=1.25in,clip,keepaspectratio]{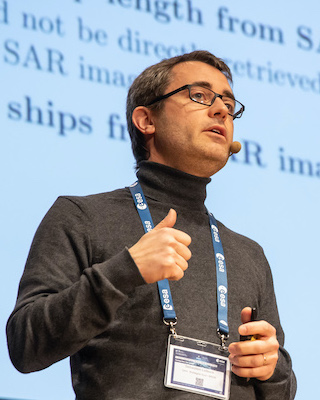}}]{Sébastien Lefèvre}
(M.Sc 1999, PhD 2002, Hab. 2009) is a Full Professor in Computer Science at Université Bretagne Sud since 2010. He is involved in numerous activities related to Artificial Intelligence for Earth and Environment Observation: founder of the OBELIX team (www.irisa.fr/obelix) within the Institute for Research in Computer Science and Random Systems (IRISA), chair of the GeoData Science track of the EMJMD Copernicus Master in Digital Earth (www.master-cde.eu), co-chair of the ECML-PKDD MACLEAN workshop series on Machine Learning for Earth Observation, etc. He is also an IEEE Senior Member and serves as an Associate Editor for IEEE TGRS.
\end{IEEEbiography}

\begin{IEEEbiography}[{\includegraphics[width=1in,height=1.25in,clip,keepaspectratio]{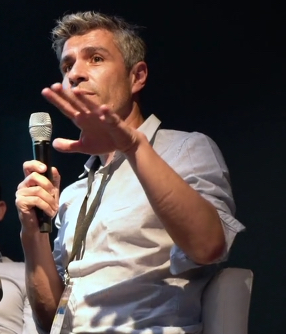}}]{Thomas Corpetti} received the Engineering degree in electrical engineering and the master’s degree in computer vision, in 1999, and the Ph.D. and Habilitation degrees in computer vision and applied mathematics with University Rennes I, France, in 2002 and 2011, respectively.  He obtained a Permanent Researcher position at the French National Institute for Scientific Research (CNRS), in 2004, on the analysis of remote sensing image sequences for environmental applications. From 2009 to 2012, he was with LIAMA, a sino-french laboratory in computer sciences, automatics, and applied mathematics at Beijing, China. He is currently with the Observatory for Universe Sciences of Rennes (OSUR), France and Littoral, Environnement, Télédétection, Géomatique (LETG) UMR 6554 as the Director of research, CNRS. His main research interest includes the definition of computer vision tools for the analysis of remote sensing data (low and high resolution) for environmental applications.
\end{IEEEbiography}

\vfill

\end{document}